\documentclass[article]{Definitions/preprint} 

\usepackage{epstopdf}
\usepackage{caption}
\usepackage{subcaption}
\usepackage{dsfont}
\usepackage{xurl}
\usepackage{rotating}
\usepackage{longtable}
\usepackage{tabularx}
\usepackage[normalem]{ulem} 
\def\UrlBreaks{\do\/\do_}

\usepackage{amssymb}
\usepackage{pifont}
\newcommand{\cmark}{\ding{51}}%
\newcommand{\xmark}{\ding{55}}%

\DeclareMathOperator*{\argmin}{arg\,min}
\DeclareMathOperator*{\argmax}{arg\,max}

\title{Is it worth it? Comparing six deep and classical methods for unsupervised anomaly detection in time series}

\author{ 
   \href{https://orcid.org/0000-0003-2264-9495}{\includegraphics[scale=0.06]{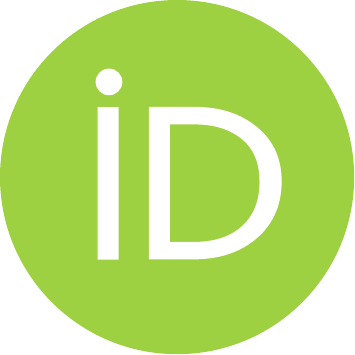}\hspace{1mm}Ferdinand Rewicki}
        \\
    Institute of Data Science\\
	German Aerospace Center\\
	07745 Jena, Germany \\
	\texttt{ferdinand.rewicki@dlr.de} \\
	\And
	\href{https://orcid.org/0000-0002-3193-3300}{\includegraphics[scale=0.06]{Definitions/logo-orcid.pdf}\hspace{1mm}Joachim Denzler} \\
	Institute of Computer Science\\
	Friedrich-Schiller University\\
	07743 Jena, Germany \\
	\texttt{joachim.denzler@uni-jena.de} \\
	\And
	\href{https://orcid.org/0000-0001-5413-2234}{\includegraphics[scale=0.06]{Definitions/logo-orcid.pdf}\hspace{1mm}Julia Niebling} \\
	Institute of Data Science\\
	German Aerospace Center\\
	07745 Jena, Germany \\
	\texttt{julia.niebling@dlr.de} \\
}

\renewcommand{\shorttitle}{Is It Worth It?}

\hypersetup{
pdftitle={Is it worth it? An experimental comparison of six deep- and classical machine learning methods for unsupervised anomaly detection in time series.},
pdfsubject={machine-learning,anomaly-detection,time-series},
pdfauthor={Ferdinand Rewicki, Joachim Denzler, Julia Niebling},
pdfkeywords={Anomaly Detection, Time Series, Machine Learning, Deep Learning, Benchmark},
}

\begin{document}
\maketitle

\begin{abstract}
Detecting anomalies in time series data is important in a variety of fields, including system monitoring, healthcare, and cybersecurity. While the abundance of available methods makes it difficult to choose the most appropriate method for a given application, each method has its strengths in detecting certain types of anomalies. In this study, we compare six unsupervised anomaly detection methods of varying complexity to determine whether more complex methods generally perform better and if certain methods are better suited to certain types of anomalies. We evaluated the methods using the UCR anomaly archive, a recent benchmark dataset for anomaly detection. We analyzed the results on a dataset and anomaly type level after adjusting the necessary hyperparameters for each method. Additionally, we assessed the ability of each method to incorporate prior knowledge about anomalies and examined the differences between point-wise and sequence-wise features. Our experiments show that classical machine learning methods generally outperform deep learning methods across a range of anomaly types.
\end{abstract}

\keywords{Anomaly Detection \and Time Series \and Machine Learning \and Deep Learning \and Benchmark}

\section{Introduction}
The detection of anomalies, or observations that significantly deviate from what is considered normal \cite{Ruff2020}, in time series data is essential in various fields, including healthcare \cite{Sabic2021}, cybersecurity \cite{Li2019, Buczak2016}, industry \cite{Sarda2021}, and robotics \cite{Park2018}. 
Anomaly detection is a notoriously challenging task, as the definition of what is considered anomalous can vary based on the context or application \cite{Freeman2022}. 
Moreover, the absence of labeled training data for non-academic problems often precludes the use of supervised machine learning techniques. 
Anomaly detection in data streams, which requires rapid results while aiming to detect anomalies accurately and efficiently, is frequently necessary. 
It is important to minimize false positive detections to prevent alarm fatigue, which can result in a serious problem being overlooked due to excessive false alarms \cite{Freeman2022}. 
It is also necessary to choose the appropriate method based on the application and, often, domain knowledge, as the existence of a universal anomaly detection method is a myth \cite{Laptev2015Generic}.
Choosing the appropriate method from the plethora of available options can be a challenge in itself, as different methods have different strengths in detecting certain types of anomalies. 
The numerous available methods can be categorized using various criteria, such as the underlying probabilistic, classification, or reconstruction-based model \cite{Ruff2020}, the type of input data (univariate or multivariate), the need for labeled training data, or the ability to process data streams. 

In this work, we compare six unsupervised anomaly detection methods with varying complexities. 
Three of these methods are classical machine learning techniques\footnote{We refer to these methods as \textit{classical} methods.} while the remaining three are based on deep learning. Our central questions in this comparison are:
\begin{enumerate}
\item "Is it worthwhile to sacrifice the interpretability of classical methods for potentially superior performance of deep learning methods?"
\item "What different types of anomalies are the methods capable of detecting?"
\end{enumerate}
To address these questions, we compare the classical methods of Robust Random Cut Forest (RRCF) \cite{Guha2016}, Maximally Divergent Intervals (MDI) \cite{Barz2018}, and MERLIN \cite{Nakamura2020} to the deep learning methods of Autoencoder (AE), Graph Augmented Normalizing Flows (GANF) \cite{Dai2021}, and Transformer Networks for Anomaly Detection (TranAD) \cite{Tuli2022}. 
We evaluate these methods on the UCR Anomaly Archive \cite{Wu2020}, a new benchmark dataset for time series anomaly detection. This archive consists of 250 univariate time series from four domains: human medicine, industry, biology, and meteorology. 
To ensure a fair comparison, we carefully design our experimental setup and perform intensive hyperparameter tuning for applicable methods. 
To the best of our knowledge, this is the first work to conduct an experimental comparison of classical and deep learning methods for anomaly detection in time series.
Our key contributions are:
\begin{itemize}
    \item We conduct a comprehensive comparison of six state-of-the-art anomaly detection methods for time series data using the UCR Anomaly Archive benchmark dataset. Our comparison is carried out in a well-defined and fair benchmark environment.
    \item We enhance the UCR Anomaly Archive by annotating it with 16 distinct anomaly types, providing a more nuanced and informative benchmark.
    \item We address two crucial questions in the field of anomaly detection: (1) whether the superior performance of deep learning methods justifies the loss of interpretability of traditional methods and (2) the similarities and differences between the analyzed methods in terms of detecting different anomaly types.
    \item We examine the impact of subsequence length on the performance of the MDI and MERLIN methods, and compare point-wise to subsequence-wise features for the RRCF method.
\end{itemize}

The remainder of this paper is organized as follows: after providing an introduction to time series data and different types of anomalies in sections \ref{subsec:intro:time_series} and \ref{subsec:intro:anomalies}, respectively, we present related work in section \ref{subsec:intro:related}. 
In section \ref{subsec:methods}, we present the six anomaly detection methods, followed by a description of the UCR Anomaly Archive dataset in section \ref{subsec:dataset} and the experimental setup in section \ref{subsec:setup}. 
The results of our experiments are presented in section \ref{sec:results} and discussed in section \ref{sec:discussion}. 
Finally, we summarize our findings and provide an outlook on future work in section \ref{sec:conclusion}

\subsection{Time Series Data}
\label{subsec:intro:time_series}
Time series are sequential data that are naturally ordered by time. 
We distinguish regular and irregular time series depending on whether or not the observations are made at equidistant intervals. 
We define a time series as an ordered set of observations based on \cite{Nakamura2020}:
\begin{Definition}
The \textbf{time series} $\mathcal{T}$ with length $n \in \mathbb{N}$ is defined as the set of pairs $\mathcal{T} = \lbrace{(t_i, p_i) | t_i \leq t_{i+1}, 0 \leq i \leq n \rbrace}$ with $p_i \in \mathbb{R}^d$ being the data points having $d$ behavioural attributes and $t_i \in \mathbb{N}, i \leq n$ the timestamps a certain data point refers to. 
For $d=1$, $\mathcal{T}$ is called univariate, and for $d>1$ $\mathcal T$ is called multivariate.
\end{Definition}
Time series can be described using different characteristics, such as \textit{stationarity}, which refers to a constant mean, variance, and auto-correlation structure, \textit{seasonality} describing periodically reoccurring behavior or \textit{sampling rate}, the frequency in which observations are made \cite{Freeman2022}. 
For an in-depth analysis of these characteristics, we refer to \cite{Freeman2022}.
As time series are usually not analyzed en bloc, we define a subsequence as a contiguous subset of the time series: 
\begin{Definition}
The \textbf{subsequence} $S_{a,b} \subseteq \mathcal{X}$ of the times series $\mathcal{X}$, with length $L = b-a > 0$ is given by $S_{a,b} := \lbrace{ (t_i, p_i) | 0 \leq a \leq i \leq b \leq n \rbrace }$. 
For simplicity, we will often omit the indices and refer to an arbitrary subsequence as $S$.
\end{Definition}

\subsection{Anomalies}
\label{subsec:intro:anomalies}
There are three main types of anomalies distinguished in the literature: point anomalies, collective anomalies, and contextual anomalies \cite{Gupta2013, Goldstein2014, BlazquezGarcia2020, Ruff2020, Braei2020}. 
Point anomalies are individual data points $(t_i, p_i) \in \mathcal{T}$ that deviate significantly from all other instances, such as a fraudulent transaction among legal finance transactions \cite{Ruff2020}. 
Collective anomalies refer to whole subsequences $\tilde{S}_{a,b} \subset \mathcal{T}$ being anomalous, while the individual data points $(t_i, p_i) \in \tilde{S}_{a,b}$ would not be considered a point anomaly \cite{Goldstein2014}. 
For example, supraventricular premature beats in an electrocardiogram (ECG) are examples of collective anomalies. 
Contextual anomalies only appear anomalous depending on specific context variables. 
For instance, while an outside air temperature measurement of $28^\circ C$ during August is considered normal in Panama, it would be anomalous in Antarctica. 
In this work, we extend this classification to 16 classes by dividing the class of collective anomalies into different subclasses, such as "frequency change" or "time shift," which are described in section \ref{subsec:dataset:anomaly_types}.

\subsection{Related Work}
\label{subsec:intro:related}
While many survey and review papers on anomaly detection are available \cite{Gupta2013, BlazquezGarcia2020, Braei2020, Chalapathy2019, Pang2020, Salehi2021, Bulusu2020} there are only a few works on comparing different methods experimentally. 

\citeauthor{Freeman2022} \cite{Freeman2022} conduct an experimental comparison of twelve anomaly detection methods like Seasonal AutoRegressive Integrated Moving Average with exogenous variables (SARIMAX), Generalized Linear Model, Facebook Prophet \cite{Taylor2018}, Matrix Profile \cite{Yeh2018} or Donut \cite{Xu2018}. 
The comparison is done using a dataset compiled mainly from the Numenta benchmark \cite{Lavin2015} with a focus on different time series characteristics like seasonality and trend. 
They use the Youden Index \cite{Fluss2005} to determine a threshold for classifying anomaly scores and assess the quality of the analyzed methods using AUC ROC, Windowed-F1, and NAB Score, which is the metric used in the Numenta benchmark. 

\citeauthor{Graabaek2022} \cite{Graabaek2022} compare 15 anomaly detection methods in the context of collaborative robots. 
The analyzed methods are categorized as \textit{instance based} like k-Nearest-Neighbors and Local Outlier Factor and \textit{explicit generalization models} such as Principal Component Analysis, One-Class Support Vector Machine or Autoencoder. 
They compare these methods on a dataset collected from different tasks performed by a robotic arm by using AUC ROC and Area under Precision-Recall Curve as quality measures.

\citeauthor{Ruff2020} \cite{Ruff2020} provide a comprehensive review of classical and deep learning methods for anomaly detection. 
They group the presented methods into the three main classes \textit{Density Estimation and Probabilistic Models}, \textit{One Class Classifications}, and \textit{Reconstruction Models} and present various classical and deep learning methods from each category. 
They also give a unifying view of the anomaly detection problem by identifying specific anomaly detection modeling components to characterize the presented methods and exemplify the modeling and evaluation process on two real-world examples.

\section{Materials and Methods}
\subsection{Analyzed Methods}
\label{subsec:methods}
To perform our comparison, we selected three deep-learning and three classical machine-learning methods for unsupervised anomaly detection. 
The selection of these methods was based on various factors, including simplicity, interpretability, applicability to data streams, and the existence of useful features such as the dependency graph for GANF. 
In the following section, we will introduce the selected methods, starting with the classical ones. 
One way to categorize anomaly detection methods is based on their suitability for handling data streams, which we will refer to as "online" anomaly detection, as opposed to "offline" anomaly detection on data batches. 
A summary of the properties of the compared methods can be found in Table~\ref{tab:methods_overview}.

\begin{table}[]
\centering
\resizebox{\textwidth}{!}{%
\begin{tabular}{@{}lllllll@{}}
\toprule
                & \textbf{mechanism} & \textbf{class} & \textbf{online/offline}                                                     & \textbf{training} & \textbf{multivariate} & \textbf{anomaly score} \\ \midrule
\textbf{RRCF}   & Isolation Forest   & classical      & online                                                                      & \xmark            & \cmark                & Collusive Displacement \\
\textbf{MDI}    & Density Estimation & classical      & offline                                                                     & \xmark            & \cmark                & (KL/JS) Divergence     \\
\textbf{MERLIN} & Discord Discovery  & classical      & offline                                                                     & \xmark            & \xmark                & Discord Distance       \\
\textbf{AE}     & Reconstruction     & deep-learning  & \begin{tabular}[c]{@{}l@{}}offline training\\ online inference\end{tabular} & \cmark            & \cmark                & Reconstruction Loss    \\
\textbf{GANF}   & Density Estimation & deep-learning  & \begin{tabular}[c]{@{}l@{}}offline training\\ online inference\end{tabular} & \cmark            & \cmark                & Density                \\
\textbf{TranAD} & Reconstruction     & deep-learning  & \begin{tabular}[c]{@{}l@{}}offline training\\ online inference\end{tabular} & \cmark            & \cmark                & Reconstruction Loss    \\ \bottomrule
\end{tabular}%
}
\caption{Overview of the properties of the anomaly detection methods considered in this comparison.}
\label{tab:methods_overview}
\end{table}

\subsubsection{Robust Random Cut Forest (RRCF)}
The Robust Random Cut Forest (RRCF) \cite{Guha2016} is a modification of the well-known Isolation Forest \cite{Liu2012} methods, that extends the approach to data streams. 
Both methods work by isolating individual points from the rest of the data by recursively partitioning the data set.
This process can be represented by a binary tree structure, where each cut is represented by a pair of branches from the same node.
The average path length can then be used as an anomaly score as shorter paths indicate that a point is more likely to be anomalous  \cite{Liu2012}. 
One key difference between RRCF and Isolation Forest is that RRCF selects the next dimension to cut with a probability proportional to the range of values in that dimension, rather than selecting it uniformly at random. 
This modification is meant to avoid cutting irrelevant dimensions and reduce the number of false positives as well as to maintain a good recall \cite{Guha2016}. 
Due to its anomaly scoring function \textit{Collusive Displacement} RRCF is also robust to the presence of duplicates or near-duplicates which could else-wise lead to outlier masking \cite{Wang2020}. 
\textit{Displacement} refers to the classification of points as outliers, if they significantly decrease the model complexity when removed from the tree. 
\textit{Collusive Displacement} accounts for duplicates or near-duplicates by removing a subset of ''colluders'' $C$ alongside the point of interest $x$ and is defined as the expected change in the depth of points in a tree when removing a set $C \cup \lbrace{x\rbrace}$. 
For an exact definition of the \textit{Collusive Displacement} scoring function please refer to \cite{Guha2016}. 
As RRCF works by isolating single points one would expect its strength in finding point anomalies. 
For the detection of anomalous subsequences, an additional preprocessing step for constructing window-based features could be considered and is analyzed in Section~\ref{subsec:results:rrcf}. 
We selected RRCF mainly due to its simplicity and comprehensibility.

\subsubsection{Maximally Divergent Intervals (MDI)}
Maximally Divergent Intervals (MDI) \cite{Barz2018} is a density-based method for offline anomaly detection in multivariate, spatiotemporal data. 
In this work, we focus on purely temporal data and provide the definitions for this case only. 
For the original definitions including spatial attributes, please refer to \cite{Barz2018}. 
Given a multivariate time series $\mathcal{T}$, MDI detects anomalous subsequences by comparing the probability density $p_S$ of a subsequence $S_{a,b} \subseteq \mathcal{T}$ to the density $p_\Omega$ of the remaining part of the times series $\Omega(S) := \mathcal{T} \setminus S_{a,b}$ for all subsequences. 
The distributions are modeled using \textit{Kernel Density Estimation} or \textit{Multivariate Gaussians}. 
To measure the degree of deviation $\mathcal{D}(p_S, p_\Omega)$ between $p_S$ and $p_\Omega$, an unbiased version of the \textit{Kullback-Leibler divergence} is used. 
The most anomalous subsequence $\tilde{S}$ is found by solving the underlying optimization problem \cite{Barz2018}:

$$
  \tilde{S} := \argmax_{S \subseteq \mathcal{T}} \mathcal{D}(p_S, p_{\Omega(S)})
$$
MDI locates this most anomalous subsequence $ \tilde{S}$ by scanning over all subsequences $S \subseteq \mathcal{T}$ with a length between $L_{min}$ and $L_{max}$ and estimates the divergence $\mathcal{D}(p_S, p_{\Omega(S)}$, which is then used as the anomaly score. 
The parameters $L_{min}$ and $L_{max}$ need to be defined in advance. 
The top-$k$ anomalous subsequences are selected by ranking the subsequences by their anomaly score and selecting the top-$k$ subsequences. 
To accommodate the application to large-scale data, an interval proposal technique based on Hotelling’s $T^2$ method \cite{MacGregor1994} is employed which selects interesting subsequences based on point-wise anomaly scores instead of performing full scans over the entire time series. 
This pre-selection method is motivated by the fact, that most subsequences are uninteresting for detecting anomalies as these are rare by definition \cite{Barz2018}. 
We selected MDI mainly due to its easily interpretable approach.

\subsubsection{MERLIN}
MERLIN \cite{Nakamura2020} is a method for offline anomaly detection based on discord discovery: Given a subsequence $S$ with length $L$ starting at timestamp $p$, a matching subsequence $M$ starting at timestamp $q$ is called a non-self match to $S$ if $|p - q| \geq L$ \cite{Nakamura2020}. 
The discord $\tilde{S}$ of a time series $\mathcal{T}$ is defined as the subsequence with the largest distance $d(\tilde{S}, M_{\tilde{S}})$ from its nearest non-self match $M_{\tilde{S}}$, where $d(\cdot,\cdot)$ is the z-normalized (zero mean and unit variance) Euclidean distance. 
MERLIN is based on the discord discovery algorithm from \cite{Yankov2007}. 
A key factor in the success and efficiency of the algorithm is the selection of the hyperparameter $r$. 
This parameter should be chosen slightly less than the discord distance, $d(\tilde{S}, M_{\tilde{S}})$. 
If $r$ is chosen too large, the algorithm will fail, while if it is too small, the runtime will be excessively long.
To address this challenge, MERLIN provides a structured search procedure for determining an appropriate value for $r$ by leveraging the observation that good values of $r$ for subsequences of length $L$ are likely to be similar to good values of $r$ for subsequences of length $L-1$ \cite{Nakamura2020}. 
The maximum value of $r$ for subsequences of length $L$ is given by $r_{max}(L) = 2\sqrt{L}$ \cite{Paepe2019}.
To find an appropriate value for $r$, the algorithm begins by setting $r = r_{max}(L_{min})$, where $L_{min}$ is the smallest subsequence length being considered, and then halving $r$ until the first discord is returned. 
For subsequences of other lengths $L_{min}, \dots, L_{max}$, the previously determined values of $r$ can be used. 
We selected MERLIN as it is the method provided with the UCR anomaly archive dataset, which is the benchmark dataset for our study and will be introduced in Section~\ref{subsec:dataset}.

\subsubsection{Autoencoder (AE)}
\begin{figure}
\centering
\includegraphics[width=10.5cm]{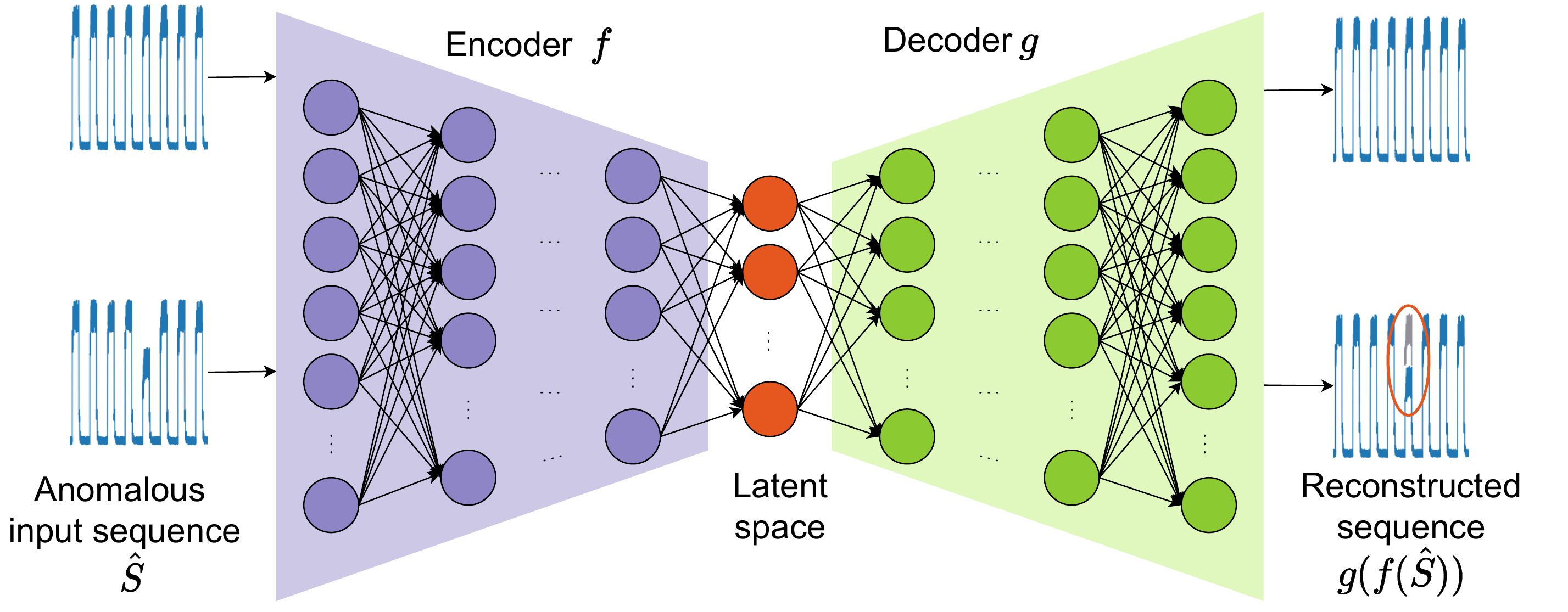}
\caption{Visualisation of a dense autoencoder model for time series anomaly detection, producing a high reconstruction error $\Delta(S, g(f(\tilde{S}))$ when presented with the anomalous subsequence $\tilde{S}$. \textit{Dense} refers to the encoder and decoder networks being fully connected. \label{fig.methods.ae}}
\end{figure}

Autoencoders, introduced in \cite{Rumelhart1987}, are neural networks designed for dimensionality reduction that consists of an encoder network $f: \mathbb{R}^n \rightarrow \mathbb{R}^l$ and a decoder network $g: \mathbb{R}^l \rightarrow \mathbb{R}^n$, where $p, n \in \mathbb{N}$ and $l < n$. 
These networks are trained to reconstruct their input by learning a latent representation. 
The autoencoder problem can be formalized according to \cite{Baldi2012} as:

$$
\argmin_{f, g} \mathbb{E}[\Delta(x, g(f(x))]
$$
with $x \in \mathcal{X}$ being the input data, $\Delta$ the reconstruction loss, i.e. usually the $L_2^2$ error function and $\mathbb{E}[\cdot]$ the expectation over its argument \cite{Baldi2012, Bank2020}. 
In the context of unsupervised anomaly detection in time series data, the autoencoder learns a normal profile of the time series $\mathcal{T}$ and detects anomalous input sequences $\tilde{S}$ by a high reconstruction error. 
Figure~\ref{fig.methods.ae} illustrates this approach.

In our experiments, we used a dense autoencoder with two hidden layers in the encoder and decoder which have each a doubled number of neurons of the latent space and ReLU activations as nonlinearities. 
We included the autoencoder as a basic deep-learning model in this study.

\subsubsection{Graph Augmented Normalizing Flows (GANF)}
GANF \cite{Dai2021} is an anomaly detection method for multivariate time series that uses normalizing flows for density estimation.
Normalizing flows are generative models $f: \mathbb{R}^d \rightarrow \mathbb{R}^d$ that utilize a series of invertible and differentiable transformations to normalize complex data distributions to "base" distributions, whose densities are typically easy to evaluate (e.g., isotropic Gaussians) \cite{Dai2021}. 
In addition to modeling the density of the time series using a normalizing flow, GANF incorporates a Bayesian Network to model the causal relationships among multiple multivariate time series $\mathcal{X} = (\mathcal{T}_1, \dots, \mathcal{T}_m)$. 
Given a training set $\mathcal{D} = \lbrace{\mathcal{X}i\rbrace{i=1}^{|\mathcal{D}|}}$ of multiple time series, GANF aims to learn the adjacency matrix $A$ of the Bayesian Network and simultaneously the graph-augmented normalizing flow $\mathcal{F}: (\mathcal{X}, A) \rightarrow \mathcal{Z}$, where $\mathcal{Z}$ is a random variable with a "simple" (base) distribution \cite{Dai2021}. 
Once $\mathcal{F}$ is learned, the estimated density $p({\mathcal{X})}$ can be evaluated to identify anomalies in low-density regions of the base distribution. 
The dependency encoder of the model consists of a recurrent neural network to summarize the time series up to a given time step $t$ and a graph convolution layer to learn a dependency representation, which is then used to condition a normalizing flow $f$. 
For more information on the architectural details of GANF, please see \cite{Dai2021}. 
As anomalies are rare by definition, it is typically assumed that their densities are low, and thus the estimated densities can be used as an anomaly score \cite{Dai2021}. 
We included GANF as a deep learning variant of a density estimation-based anomaly detection method, given its ability to learn dependencies between multiple time series, although this feature is not used in the context of this comparison.

\subsubsection{Transformer Network for Anomaly Detection (TranAD)}
TranAD \cite{Tuli2022} is an anomaly detection method based on the Transformer model \cite{Vaswani2017}, which learns to reconstruct an input by applying several attention-based transformations. 
The model proposed by \citeauthor{Tuli2022} \cite{Tuli2022} uses a two-phase training. 
In the first phase, the model learns an approximate reconstruction of the whole time series $\mathcal{T}$ to capture long-term trends and uses the deviation from the true time series as a \textit{focus score}. 
In phase two the focus score is used to find those subsequences where the deviation in phase one was high. 
Similar to other encoder-decoder models, the reconstruction loss is used as the anomaly score. 
We included TranAD as it was one of the most recent publications by the time of its selection. 

\subsection{Benchmark Dataset: UCR Anomaly Archive}
\label{subsec:dataset}
The dataset used in this study is the \textit{UCR Anomaly Archive} \cite{Wu2020, Ucraa2021}, which consists of 250 univariate time series from various fields including human medicine, biology, meteorology, and industry. 
The time series in this dataset include both natural and artificial anomalies, with the majority being artificial. 
This allows for a more detailed analysis based on the type of anomaly injection. 
The UCR Anomaly Archive was first used in an anomaly detection contest preceding the ACM SIGKDD conference in 2021 and was published by \citeauthor{Wu2020} \cite{Wu2020} as an alternative to commonly used benchmark datasets such as Yahoo S5 \cite{Laptev2015}, Numenta \cite{Lavin2015} or NASA \cite{Hundman2018}, which have been criticized for having trivial anomalies, unrealistic anomaly densities, mislabeled ground truth, and being "run-to-failure biased." 
This term refers to anomalies occurring at the end of a time series due to the recording being stopped after the anomaly (or \textit{failure}) occurred. 

Each time series in the UCR Anomaly Archive contains a single, sometimes subtle anomaly after a certain time stamp, with the data before that time stamp being considered normal. 
As we evaluate the methods described in Section~\ref{subsec:methods} in the unsupervised setting, we do not use this label.
The time series in the UCR Anomaly Archive have lengths ranging from 6674 to 900000 data points and anomalies with lengths between 1 and 1701 data points, with a maximum anomaly pollution of 4.9\% per time series.

\subsubsection{Included Time Series}
\begin{figure}
\centering
    \begin{subfigure}[t]{0.53\textwidth}
        \centering
        \includegraphics[width=\textwidth]{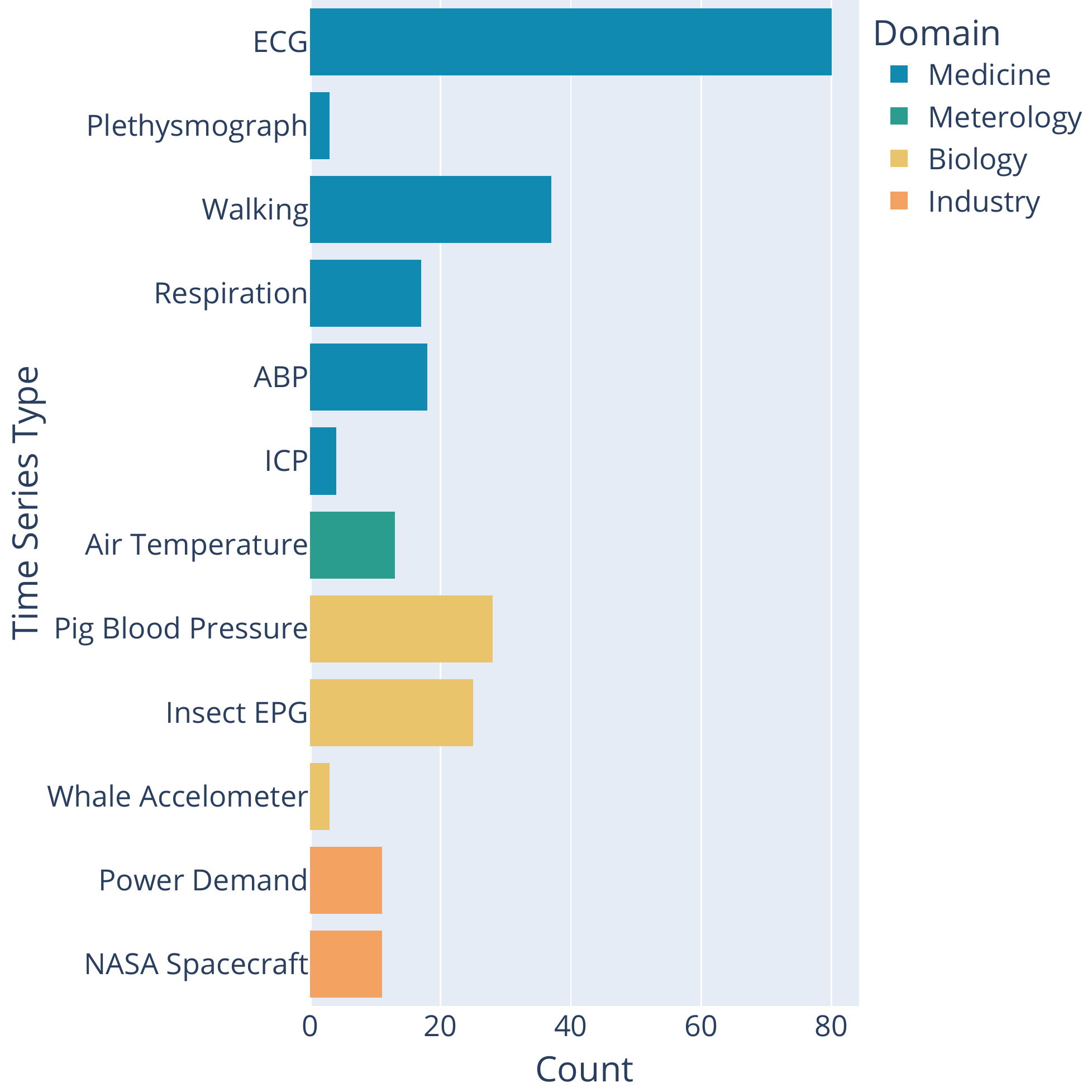}
        \caption{Histogram of time series types included in the UCR Anomaly Archive. The color indicates the domain, the time series originates from.}
        \label{fig.dataset.time_series_types}
    \end{subfigure}
    \hfill
    \begin{subfigure}[t]{0.45\textwidth}
        \centering
        \includegraphics[width=\textwidth]{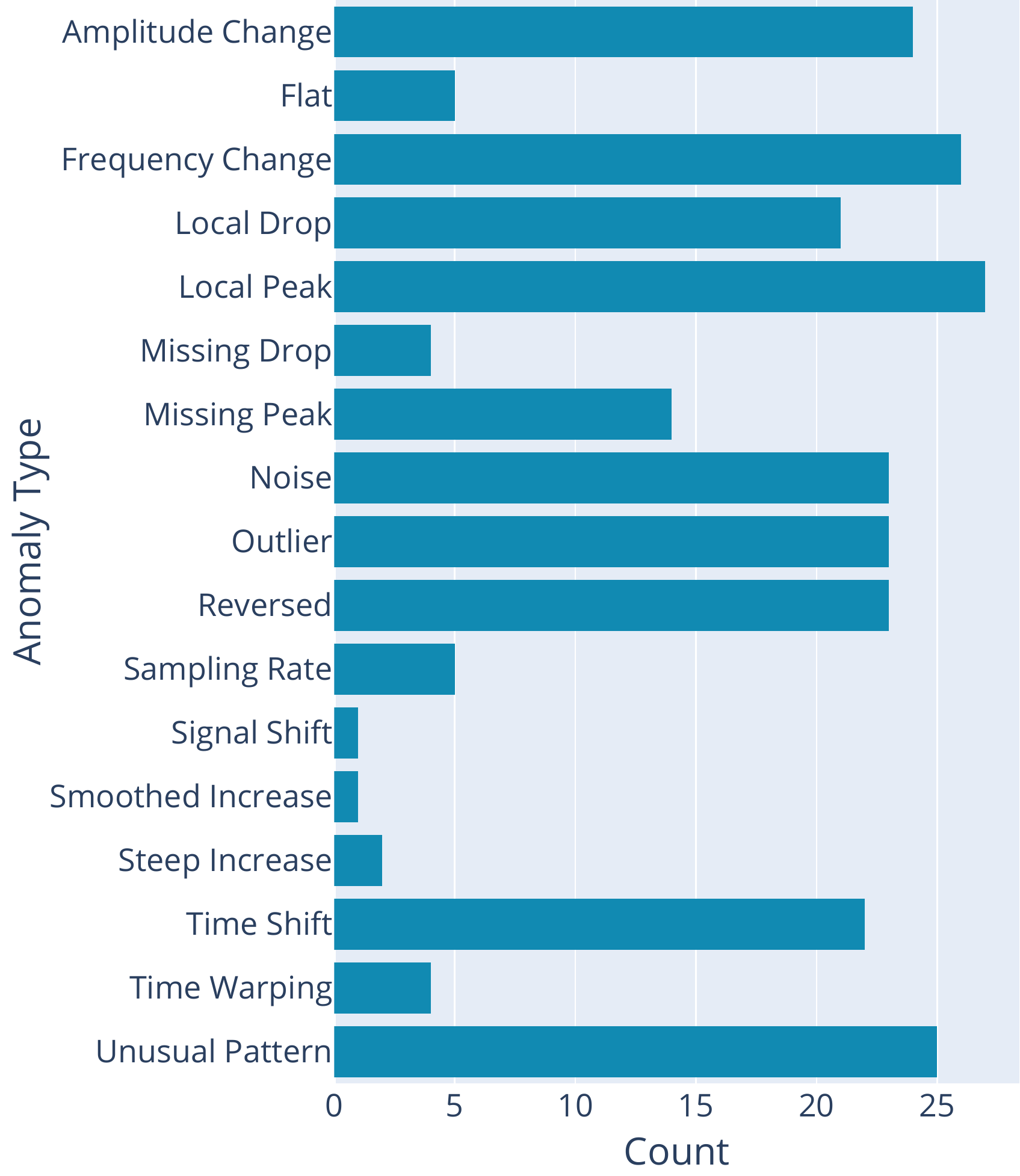}
        \caption{Histogram of anomaly types present in the UCR Anomaly Archive.}
        \label{fig.dataset.anomaly_types}
    \end{subfigure}
    \caption{}
    \label{fig:dataset:histograms}
\end{figure}

The time series in the UCR Anomaly Archive can be classified into 12 types based on the domain they originate from: human medicine, meteorology, biology, and industry. Figure~\ref{fig.dataset.time_series_types} shows the distribution of time series types in the dataset.

The distribution is highly imbalanced, with approximately 64\% of the time series coming from human medicine applications, 22\% from biology, 9\% from industry, and 5\% being air temperature measurements. Within a single type of time series (e.g., ECG), the time series are not unique but differ in terms of injected anomalies or modifications to the original time series, such as the addition of Gaussian noise or baseline wander. 
Baseline wander, a low-frequency artifact commonly found in ECG caused by factors such as breathing or subject movement, refers to slow changes in the signal baseline \cite{Lenis2017}.

\subsubsection{Anomaly Types}
\label{subsec:dataset:anomaly_types}
To evaluate the abilities of the six anomaly detection methods to detect different types of anomalies, we annotated each time series with the type of injected anomaly. 
We used the supplemental material provided with the UCR Anomaly Archive dataset \cite{Ucraa2021Supl} to obtain this information. 
The distribution of anomaly types is shown in Figure~\ref{fig.dataset.anomaly_types}. 
A list of explanations and examples for each anomaly type can be found in Appendix~\ref{appendix:anomaly_types}.

\subsection{Experimental Setup}
\label{subsec:setup}
For our experiments, we implemented the benchmark pipeline described in Section~\ref{subsec:pipeline} in Python.
The anomaly detection methods were either integrated from their publicly available GitHub repositories (RRCF\footnote{\url{https://github.com/kLabUM/rrcf}}, MDI\footnote{\url{https://github.com/cvjena/libmaxdiv}}, GANF\footnote{\url{https://github.com/EnyanDai/GANF}}, TranAD\footnote{\url{https://github.com/imperial-qore/TranAD}}) or implemented by us if a Python version was not available (MERLIN\footnote{\url{https://gitlab.com/dlr-dw/py-merlin}}). 
The Autoencoder model was implemented using the PyTorch\cite{Pytorch2019} library and is available in the repository for this work\footnote{\url{https://gitlab.com/dlr-dw/is-it-worth-it-benchmark}}. 
The relevant hyperparameters for each model were tuned through 20 rounds of Bayesian optimization on 25 randomly selected time series from the UCR Anomaly Archive, using the F1 score as the optimization target.
The time series used for hyperparameter tuning were excluded from the actual experiments. A table containing all hyperparameters obtained from that search can be found in Appendix~\ref{appendix:parameters} 
All experiments were run on an Intel Xeon Platinum 8260 CPU with 10GB of allocated memory\footnote{For TranAD, we increased the memory to 20GB for the timeseries\\''239\_UCR\_Anomaly\_taichidbS0715Master\_190037\_593450\_593514.txt'',\\ ''240\_UCR\_Anomaly\_taichidbS0715Master\_240030\_884100\_884200.txt'' and \\ ''241\_UCR\_Anomaly\_taichidbS0715Master\_250000\_837400\_839100.txt''.}. 
We ran all experiments six times: the first time we set the random number generators of Python, Numpy, and PyTorch to a fixed value\footnote{We used 42 as the seed value across all experiments.} and then performed 5 repetitions without setting a random seed to account for random sampling effects.

\subsubsection{Benchmark Pipeline}
\label{subsec:pipeline}
\begin{figure}[H]
\centering
\includegraphics[width=12.5cm]{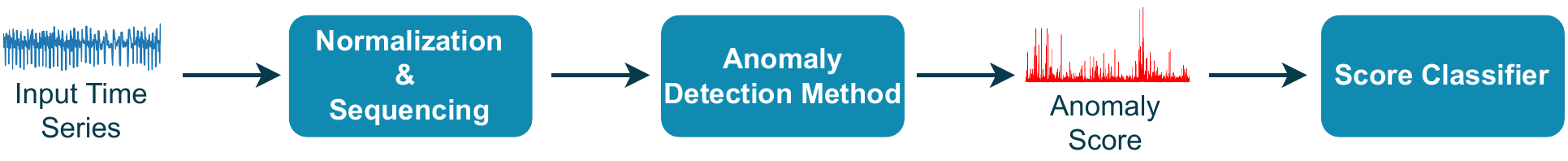}
\caption{Benchmark pipeline used in the experiments. \label{fig.benchmark.pipeline}}
\end{figure}

To maintain a controlled experimental environment and ensure fairness among all experiments, we implemented the pipeline shown in Figure~\ref{fig.benchmark.pipeline}. 
The time series data were normalized to the interval $[0,1]$ and a sliding window approach was applied, depending on the requirements of each method. 
While AE, GANF, and TranAD require input data to be given as subsequences with fixed length $L$, MDI and MERLIN require the entire time series along with a range of subsequence lengths $L_{min}$ and $L_{max}$. 

In the case of the MDI and MERLIN methods, the range of subsequence lengths used was arbitrarily set to $L_{min} = 75$ and $L_{max} = 125$ time steps. For TranAD, the subsequence length $L = 10$ and stride $s = 1$ were used, according to the experiments in \cite{Tuli2022} for the time series taken from the UCR anomaly archive. For GANF, a subsequence length of $L = 100$ was chosen, which is the middle of the range $L_{min}$ and $L_{max}$, and stride $s = 10$ based on \cite{Dai2021}. As for the AE method, the subsequence length $L = 10$ and stride $L = 10$ were determined empirically. RRCF does not require the data to be given as subsequences. A table, summarizing the configurations used in our experiments can be found in Appendix~\ref{appendix:parameters}.

The normalized time series or subsequences were then used as input for the respective anomaly detection method, which calculates an anomaly score. 
MERLIN is an exception in this regard, as it returns only the anomalous subsequences.
As the scores produced by the different methods are very heterogeneous, we employed a method called \textit{Peak Over Threshold (POT)} \cite{Siffer2017} to determine a suitable threshold for classifying subsequences as normal or anomalous. 
This approach was also used in previous works such as \cite{Hundman2018, Boniol2020, Tuli2022}.

\subsubsection{Anomaly Score Classification}
To ensure a fair comparison of the results produced by the six anomaly detection methods, we use the principle of Extreme Value Theory (EVT) to determine a threshold for classifying subsequences as anomalous or normal \cite{Siffer2017}. 
EVT is an approach for finding the law of extreme values, which are often located in the tails of a probability distribution, without making any assumptions about the data distribution \cite{Su2019}. 
The \textit{Peaks-Over-Threshold (POT)} method \cite{Siffer2017}, which is the second theorem in EVT, fits the tail of a probability distribution with a Generalized Pareto Distribution (GPD).

In the context of anomaly score classification, the Peak Over Threshold (POT) method is utilized to learn an appropriate threshold for the anomaly scores $\lbrace{\sigma_1, \sigma_2, \dots, \sigma_n \rbrace}$ \cite{Su2019}. 
Specifically, the Generalized Pareto Distribution (GPD) is adapted to focus on values at the low ends of the distribution.
According to \cite{Su2019}, a modified version of POT for anomaly score classification is defined as follows:
Given a random variable $S$ that models the anomaly scores $\lbrace{\sigma_1, \sigma_2, \dots, \sigma_n \rbrace}$ and an initial threshold $th^0$, the cumulative distribution function of the GPD is adapted to:
\begin{equation}\label{eq:pot:cdf}
\tilde{F}(s) = \mathbb{P}(th^0 - S > s | S < th^0) \sim (1 + \frac{\gamma s}{\beta})^{-\frac{1}{\gamma}} \text{ ,}
\end{equation}
where $\beta$ and $\gamma$ are the scale and shape parameters of the GPD.
The threshold $th$ is then computed by:
\begin{equation}\label{eq:pot:th}
th \simeq th^0 - \frac{\hat{\beta}}{\hat{\gamma}}((\frac{qn}{n_{th^0}})^{-\hat{\gamma}} - 1) \text{ ,}
\end{equation}
where $\hat{\beta}$ and $\hat{\gamma}$ are the maximum likelihood estimates of the scale and shape parameters in Equation~\ref{eq:pot:cdf} estimated from $\lbrace{\sigma_1, \sigma_2, \dots, \sigma_n \rbrace}$, $q$ is the preferred probability to observe an anomaly score below the initial threshold $S < th^0$ and $n_{th^0}$ is the number of anomaly scores below the initial threshold $|\lbrace{\sigma_i | \sigma_i < th^0\rbrace}|$.
The anomaly label $y_i$ for a predicted subsequence $\hat{S}_{i,i+w}$ of length $w$ is obtained by:
\begin{equation*}
y_i = \mathds{1}(\sigma_i \geq th) \text{ ,}
\end{equation*}
where $\sigma_i$ is the anomaly score for the subsequence $\hat{S}_{i,i+L}$, $th$ is the threshold from Equation~\ref{eq:pot:th}, and $\mathds{1}(\cdot)$ is the indicator function.
For further information on POT, interested readers may refer to \cite{Su2019} and \cite{Siffer2017}. 
In the experiments, the Streaming POT variant from \cite{Siffer2017} is used, with POT being initialized on the first 10\% of the anomaly scores and the parameter $q$ set to $0.01$ empirically.

We do not perform this step for the MERLIN method because it already returns binary labels per subsequence. 
Since it is typically acceptable for an algorithm to detect any point in an anomalous subsequence as long as the delay is not too long, we adopt the method proposed in \cite{Xu2018} and subsequently used in \cite{Zhao2020, Tuli2022} for adjusting the predicted anomalous labels to account for varying subsequence lengths. 
If a point in a true anomalous segment can be detected by the derived score and threshold, we count this segment as correctly detected from that point forward and treat all points within the segment as if they could be detected by the threshold.

\subsubsection{Quality Measures}
\label{sec:methods:measures}
To evaluate the performance of the anomaly detection methods, we use the area under the receiver operating characteristic curve (AUC ROC), F1 Score, and UCR score. The UCR score\footnote{The scoring function is not named in \cite{Ucraa2021Supl}, so we call it UCR score} is the recommended metric provided with the UCR Anomaly Archive.
To calculate the AUC ROC and F1 Score, we scale the anomaly scores for the subsequences back to the length of the subsequence and calculate point-wise metrics.

\paragraph{AUC ROC}
The AUC ROC is a measure of the ability of a binary classifier to separate two classes and can be seen as a single-number summary of a ROC plot \cite{Bradley1997}. 
In a ROC plot, the true positive rate is plotted against the false positive rate at increasing threshold levels for thresholding the output of the classifier.
The higher the AUC ROC, the better the classifier can separate the two classes. 
A perfect classifier achieves a score of 1 by ranking all examples of the positive class higher than all examples of the negative class. 
Therefore, we use the AUC ROC as a measure of the quality of the produced anomaly scores, where a high score indicates good separability between normal and anomalous points or subsequences. 
We are aware that the AUC ROC is not a suitable measure for unbalanced problems such as anomaly detection, where the anomalous class is small by definition, but we report it due to its widespread use in the literature.

\paragraph{F1 Score}
The F1-Score is the harmonic mean of precision and recall and is defined as
$$
F1 = 2 \cdot \frac{precision \cdot recall}{precision + recall} = \frac{TP}{TP + \frac{1}{2}(FP+FN)}
$$
where TP, FP, and FN are the True Positive, False Positive, and False Negative detections, and precision and recall are defined as:
$$
precision = \frac{TP}{TP + FP} \text{, } recall = \frac{TP}{TP + FN} \text{   .}
$$ 
Since the F1 Score is calculated based on the result of the binary classification, it provides evidence about the quality of the threshold used. If a method has a high AUC ROC but a low F1 Score, this would indicate a poor threshold.

\paragraph{UCR score}
The UCR score is the recommended metric provided with the UCR Anomaly Archive and is a binary score indicating whether or not a method was able to find the single anomaly in a time series.
It is defined as:
\begin{equation}
\label{eq:ucrscore}
    UCR_{\text{score}} := \mathds{1}(min(a - L_{\tilde{S}}, a - 100) < t^* < max(b + L_{\tilde{S}}, b + 100))
\end{equation}
where $a$ and $b$ are the beginning and end of the true anomaly with length $L_{\tilde{S}}$, $t^*$ is the timestamp of the point with the highest anomaly score, and $\mathds{1}(\cdot)$ is the indicator function. 
For subsequences $S_{i,j}$ we use the middle point $t^* = i + \lfloor \frac{j-i}{2} \rfloor$. 
The tolerance of 100 time steps is added to account for very short anomalies \cite{Ucraa2021Supl}. 
Being a binary measure, the UCR score tells whether or not the single anomaly in a certain time series was detected by having the highest anomaly score. 
However, a UCR score of zero does not convey any information about whether the anomaly was found, but a false positive result has a higher anomaly score or was not detected at all. 
As this might not be necessary for a situation like a challenge, where only positive results matter, it is essential to consider other metrics like the F1 score alongside visual inspection to correctly interpret the results. 
If a method shows a UCR score of 1 but a low F1 score at the same time, it indicates the detection of the true anomaly with the highest anomaly score, as otherwise, the UCR score would be zero.
The low F1 score, however, can be either caused by the presence of false positive or false negative results or it is subject to the detection of a short anomaly within the 100 time steps tolerance interval considered in Equation~\ref{eq:ucrscore}. 
Therefore, evaluating the F1 score alone is not sufficient.
On the other hand, if a method shows a UCR score of 0 but a high F1 score, it indicates that the anomaly was identified without many false positives or false negatives, but that the anomaly score for the true anomaly ranked lower than for false detections. 
The reasons for false positives or false negatives can be manifold and will be discussed in more detail in the beginning of Section~\ref{sec:discussion}.
When interpreting aggregated UCR scores, an averaged UCR score of 0.5 means that the true anomaly was successfully identified as having the highest anomaly score in half of the analyzed time series.

\section{Results}
\label{sec:results}
We analyze the six anomaly detection methods regarding their overall performance in Section~\ref{subsec:results:by_model} and their differences in detecting certain types of anomalies in Section~\ref{subsec:results:by_anomaly_type}. 
Beyond that, we analyze the influence of varying subsequence length on MDI and MERLIN and thus their ability to utilize additional information about the anomalies in Section~\ref{subsec:results:mdi_merlin} and compare the point-wise application of RRCF to the raw time series to that on subsequence-wise statistical vectors in Section~\ref{subsec:results:rrcf}.

\subsection{Performance Analysis by Method}
\label{subsec:results:by_model}
\begin{figure}
\centering
\begin{subfigure}[t]{0.57\textwidth}
         \centering
         \includegraphics[width=\textwidth]{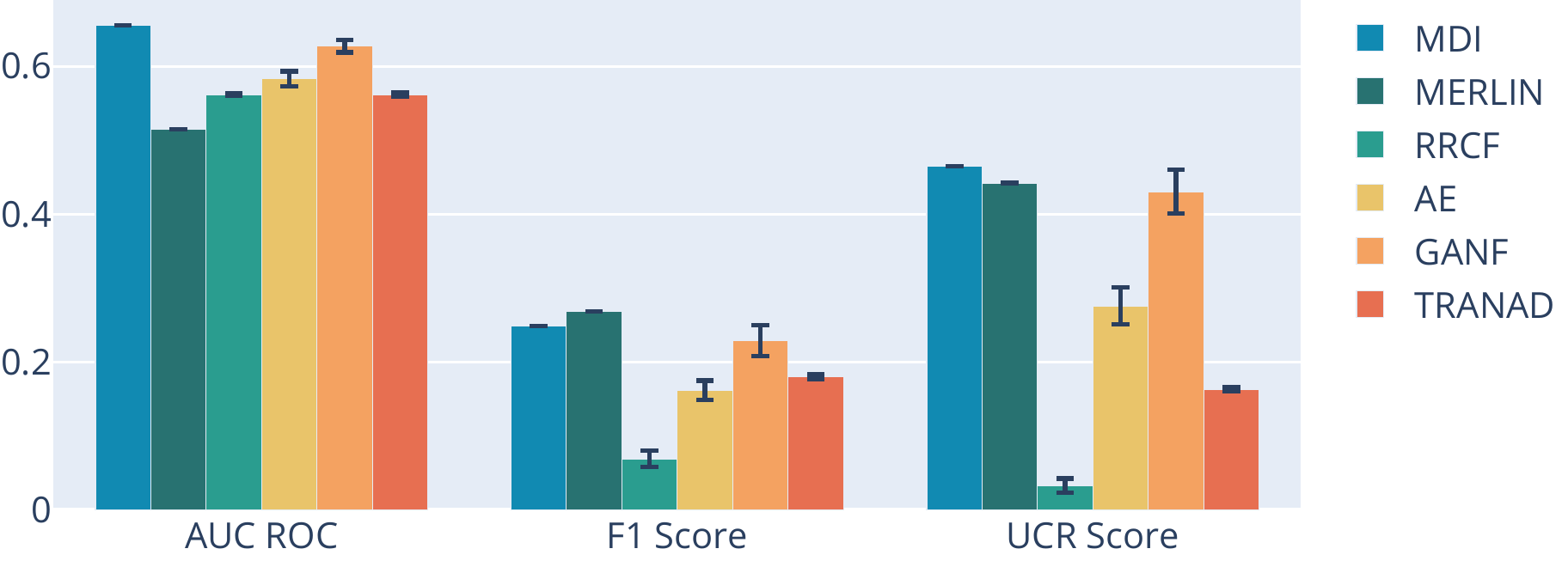}
         \caption{Macro-averaged performance metrics for each method. The error bars indicate the standard deviation caused by random effects over six repetitions of the experiment.}
         \label{fig.results.metrics_by_model}
     \end{subfigure}
     \hfill
     \begin{subfigure}[t]{0.41\textwidth}
         \centering
         \includegraphics[width=\textwidth]{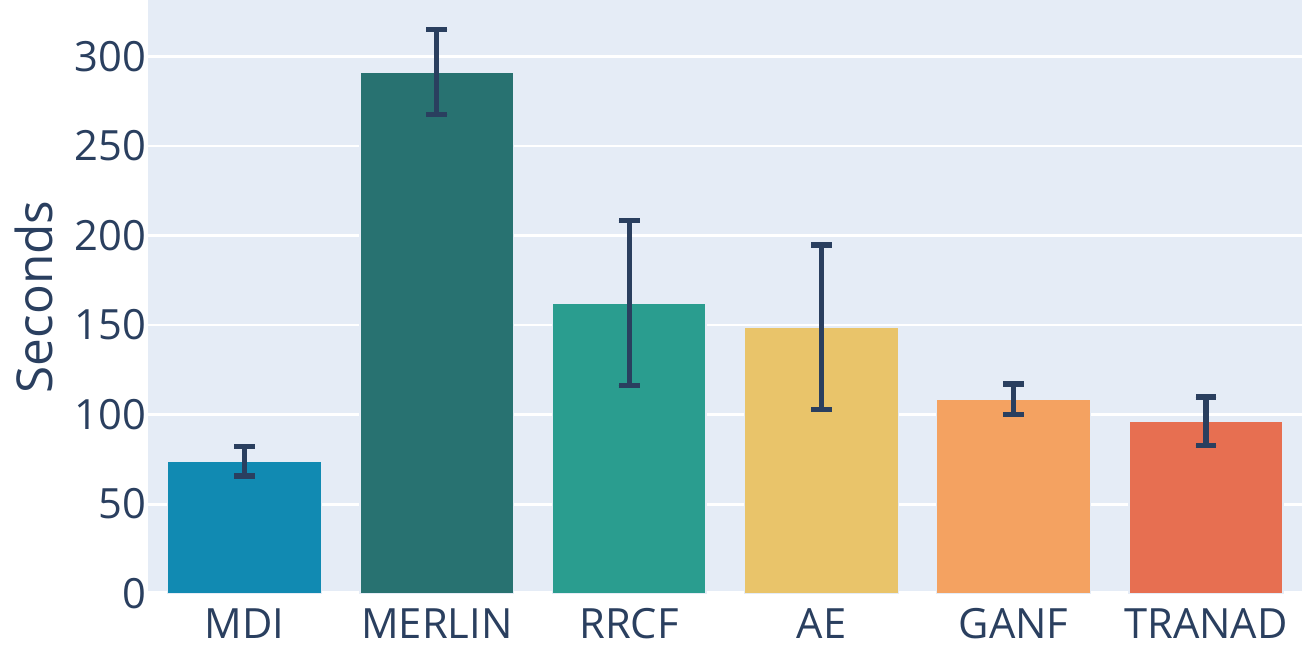}
         \caption{Average runtime per time series. For the deep-learning methods, training time is included. The error bars indicate the standard deviation over six repetitions of the experiment.}
         \label{fig.results.runtime_by_model}
     \end{subfigure}
     \caption{\label{fig.results.by_model}}
\end{figure}

We evaluate the performance of six anomaly detection methods using three metrics: macro-averaged AUC ROC, F1 score, and UCR score, as well as the average runtime for a single time series.
The results are visualised in Figure~\ref{fig.results.by_model}.
Of the methods compared, MDI achieves the highest AUC ROC and UCR scores, while MERLIN performs better in terms of F1 score.
Among the deep learning methods, GANF has the highest scores across all three metrics. F1- and UCR scores are 4\% less compared to the best-performing classical method and the AUC ROC of $0.66$ is second-best.
AE scores higher than TranAD for AUC ROC and UCR Score, while TranAD shows a slightly higher F1 score.

RRCF performs poorly, failing to detect a notable amount of anomalies in the test set and having the lowest F1 and UCR scores.
The numerical results are shown in Table~\ref{tab:results_by_method}.
The scores in Table~\ref{tab:results_by_method} are generally low across all methods, likely due to the test set time series producing low or zero scores.
We discuss the implications of various combinations of high and low scores, as well as their potential causes, in Section~\ref{sec:discussion}.
MDI and MERLIN, being deterministic methods, are not subject to sampling effects and therefore have a standard deviation of $0.0$ among the six repetitions of the experiment.

In terms of average runtime for a single time series from the UCR anomaly Archive, MDI performs best with a runtime of 74 seconds.
TranAD is about 22 seconds slower on average (96s) and GANF has an average runtime of 109 seconds.
AE has a runtime that is twice as long as that of MDI, while RRCF (162s) has a slightly longer runtime, but both fall below 200 seconds.
MERLIN has the worst runtime of 291 seconds, almost four times that of MDI.
It is worth noting that these runtimes may be influenced by the specific implementations used.
MDI is implemented in C++ with a Python interface, while the other methods are purely implemented in Python.
For the deep learning methods AE, GANF, and TranAD, the training time is included in the reported runtime.

\begin{table}
\centering
\resizebox{\textwidth}{!}{%
\begin{tabular}{llllllllr}
\hline
\multicolumn{1}{|l|}{Class}                          & \multicolumn{1}{l|}{Method} & \multicolumn{2}{c|}{AUC ROC}                                                                                       & \multicolumn{2}{c|}{F1 Score}                                                                                  & \multicolumn{2}{c|}{UCR score}                                                                                    & \multicolumn{1}{l|}{runtime (sec)}                 \\ \hline
\multicolumn{1}{|l|}{\multirow{3}{*}{Classical ML}}  & \multicolumn{1}{l|}{MDI}    & \multicolumn{1}{l|}{\textbf{0.66 $\pm 0.0$}} & \multicolumn{1}{l|}{\multirow{3}{*}{0.58 $\pm 0.0006$}} & \multicolumn{1}{l|}{0.25 $\pm 0.0$}       & \multicolumn{1}{l|}{\multirow{3}{*}{\textbf{0.20 $\pm 0.004$}}} & \multicolumn{1}{l|}{\textbf{0.47 $\pm 0.0$}} & \multicolumn{1}{l|}{\multirow{3}{*}{\textbf{0.31 $\pm 0.0033$}}} & \multicolumn{1}{r|}{\textbf{74}} \\
\multicolumn{1}{|l|}{}                               & \multicolumn{1}{l|}{MERLIN} & \multicolumn{1}{l|}{0.51 $\pm 0.0$}             & \multicolumn{1}{l|}{}                                            & \multicolumn{1}{l|}{\textbf{0.27 $\pm 0.0$}} & \multicolumn{1}{l|}{}                                           & \multicolumn{1}{l|}{0.44 $\pm 0.0$}            & \multicolumn{1}{l|}{}                                            & \multicolumn{1}{r|}{291}                           \\
\multicolumn{1}{|l|}{}                               & \multicolumn{1}{l|}{RRCF}   & \multicolumn{1}{l|}{0.56 $\pm 0.0019$}          & \multicolumn{1}{l|}{}                                            & \multicolumn{1}{l|}{0.07 $\pm 0.011$}        & \multicolumn{1}{l|}{}                                           & \multicolumn{1}{l|}{0.03 $\pm 0.0094$}         & \multicolumn{1}{l|}{}                                            & \multicolumn{1}{r|}{162}                           \\ \hline
\multicolumn{1}{|l|}{\multirow{3}{*}{Deep Learning}} & \multicolumn{1}{l|}{AE}     & \multicolumn{1}{l|}{0.58 $\pm 0.01$}            & \multicolumn{1}{l|}{\multirow{3}{*}{\textbf{0.59} $\pm 0.002$}}           & \multicolumn{1}{l|}{0.16 $\pm 0.013$}        & \multicolumn{1}{l|}{\multirow{3}{*}{0.19 $\pm 0.009$}}          & \multicolumn{1}{l|}{0.28 $\pm 0.025$}          & \multicolumn{1}{l|}{\multirow{3}{*}{0.29 $\pm 0.007$}}           & \multicolumn{1}{r|}{149}                           \\
\multicolumn{1}{|l|}{}                               & \multicolumn{1}{l|}{TranAD} & \multicolumn{1}{l|}{0.56 $\pm 0.003$}          & \multicolumn{1}{l|}{}                                            & \multicolumn{1}{l|}{0.18 $\pm 0.003$}       & \multicolumn{1}{l|}{}                                           & \multicolumn{1}{l|}{0.16 $\pm 0.003$}         & \multicolumn{1}{l|}{}                                            & \multicolumn{1}{r|}{109}                           \\
\multicolumn{1}{|l|}{}                               & \multicolumn{1}{l|}{GANF}   & \multicolumn{1}{l|}{0.63 $\pm 0.009$}           & \multicolumn{1}{l|}{}                                            & \multicolumn{1}{l|}{0.23 $\pm 0.021$}        & \multicolumn{1}{l|}{}                                           & \multicolumn{1}{l|}{0.43 $\pm 0.03$}           & \multicolumn{1}{l|}{}                                            & \multicolumn{1}{r|}{96}                           \\ \hline
                                                     &                             &                                                 &                                                                  &                                              &                                                                 &                                                &                                                                  &                                                   
\end{tabular}
}
\caption{Performance comparison of six anomaly detection methods on macro-averaged AUC ROC, F1 score, UCR score, and runtime for a single time series. 
Results are grouped by model class (classical ML and deep learning) and presented as mean $\pm$ standard deviation over six repetitions. 
The value in each second column denotes the mean aggregated by method class.}
\label{tab:results_by_method}
\end{table}

\begin{figure}
\centering
\includegraphics[width=12.5cm]{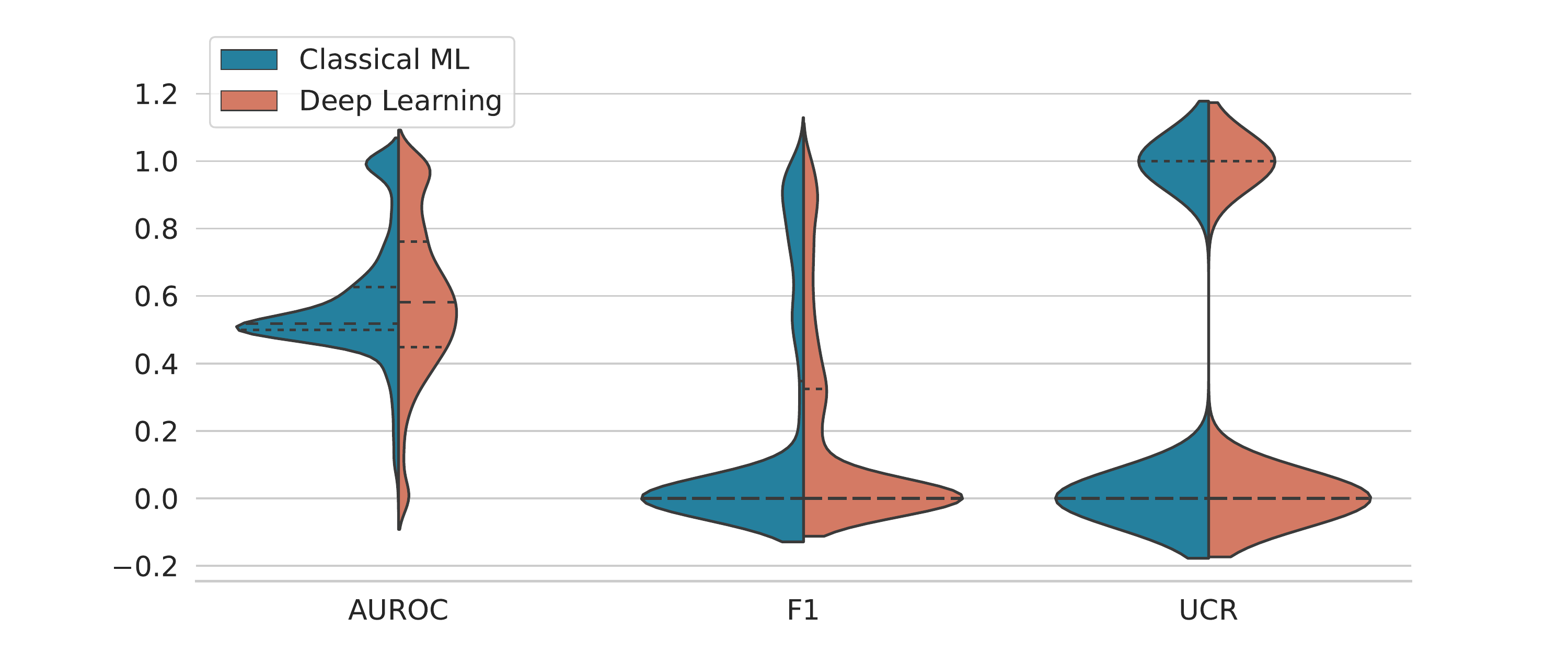}
\caption{Distribution of results for the AUC ROC, F1 Score, and UCR score aggregated by model class. The dashed lines encode the quartiles of the distributions. The area under the peaks around $1$ is larger for the Classical ML methods for F1- and UCR score.}
\label{fig.results.by_modelclass}
\end{figure}

To target the main question addressed in this paper, we aggregated the results by model class and visualized them using violin plots in Figure~\ref{fig.results.by_modelclass}.
The violin plots show the kernel density estimates for the two classes: "Classical Machine Learning Methods" (containing MDI, MERLIN, and RRCF) and "Deep Learning Methods" (containing AE, GANF, and TranAD).
All density curves have two peaks: one around 0.5 for AUC ROC and 0.0 for F1 and UCR scores, and a smaller one around 0.9 (F1 score) and 1 (AUC ROC and UCR score).
The peaks around 0.5 and 0 represent those results where the methods failed to detect anomalies, while the peaks around 0.9 and 1 mark successful anomaly detection.
For F1- and UCR score, the area under the peaks at 0.9 and 1 is larger for the classical ML methods than for the deep learning methods, indicating more successful anomaly detection for the "Classical ML" class.
Conversely, the area under the peaks at 0.5 and 0 is larger for the deep learning methods.

\subsection{Performance Analysis by Anomaly Type}
\label{subsec:results:by_anomaly_type}
\begin{figure}
\centering
\includegraphics[width=\linewidth]{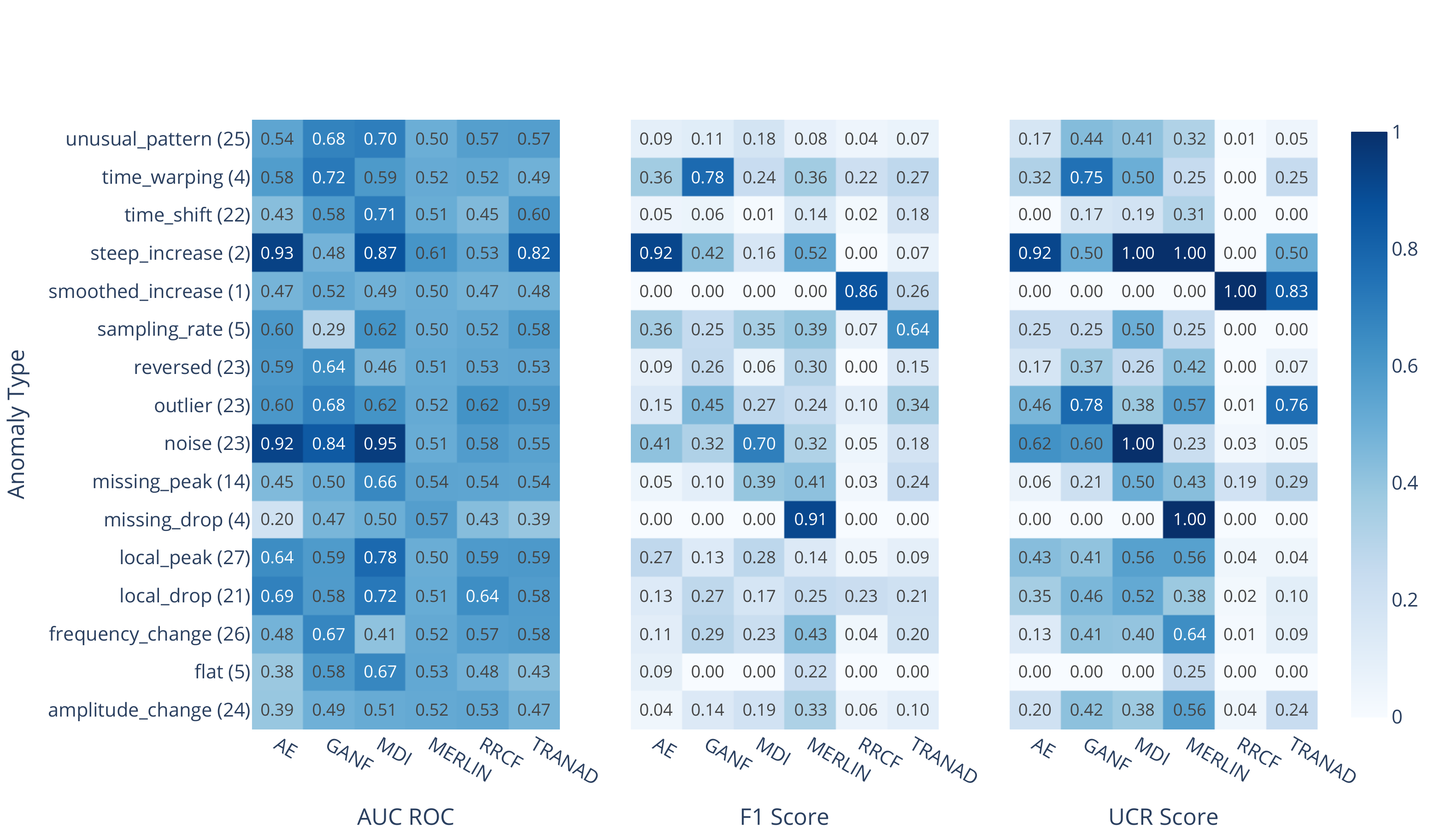}
\caption{The heatmaps show the macro-averaged AUC ROC, F1, and UCR score for the 16 annotated anomaly types over six repetitions of the experiment. 
Next to the anomaly type, the number of times series containing that type is shown in parenthesis. 
The standard deviations resulting from random effects can be found in Appendix~\ref{appendix:stds_by_anomaly_type}}
\label{fig.results.by_anomaly_type}
\end{figure}

The second interest in this study is the differences between the analyzed methods to detect certain types of anomalies. 
We, therefore, aggregated our results by the 16 anomaly classes described in Section~\ref{subsec:dataset:anomaly_types}. 
The results are shown in Figure~\ref{fig.results.by_anomaly_type}. 

\begin{figure}
\centering
    \begin{subfigure}[t]{0.47\textwidth}
        \centering
        \includegraphics[width=\textwidth]{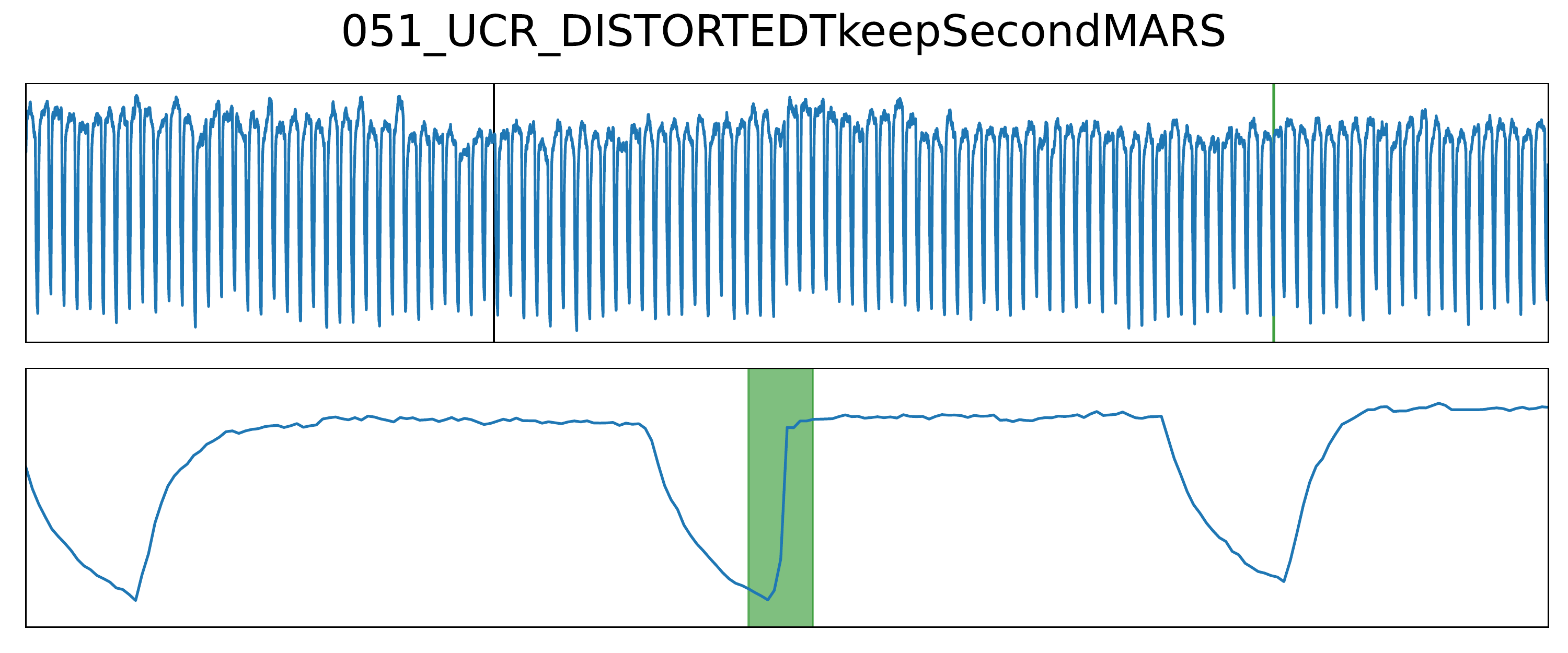}
        \caption{Overview (top) and detail (bottom) plot of the distorted time series containing the ''steep increase'' anomaly. The ground truth anomaly is highlighted in green.}
        \label{fig.disc.steep_increase}
    \end{subfigure}
    \hfill
    \begin{subfigure}[t]{0.47\textwidth}
        \centering
        \includegraphics[width=\textwidth]{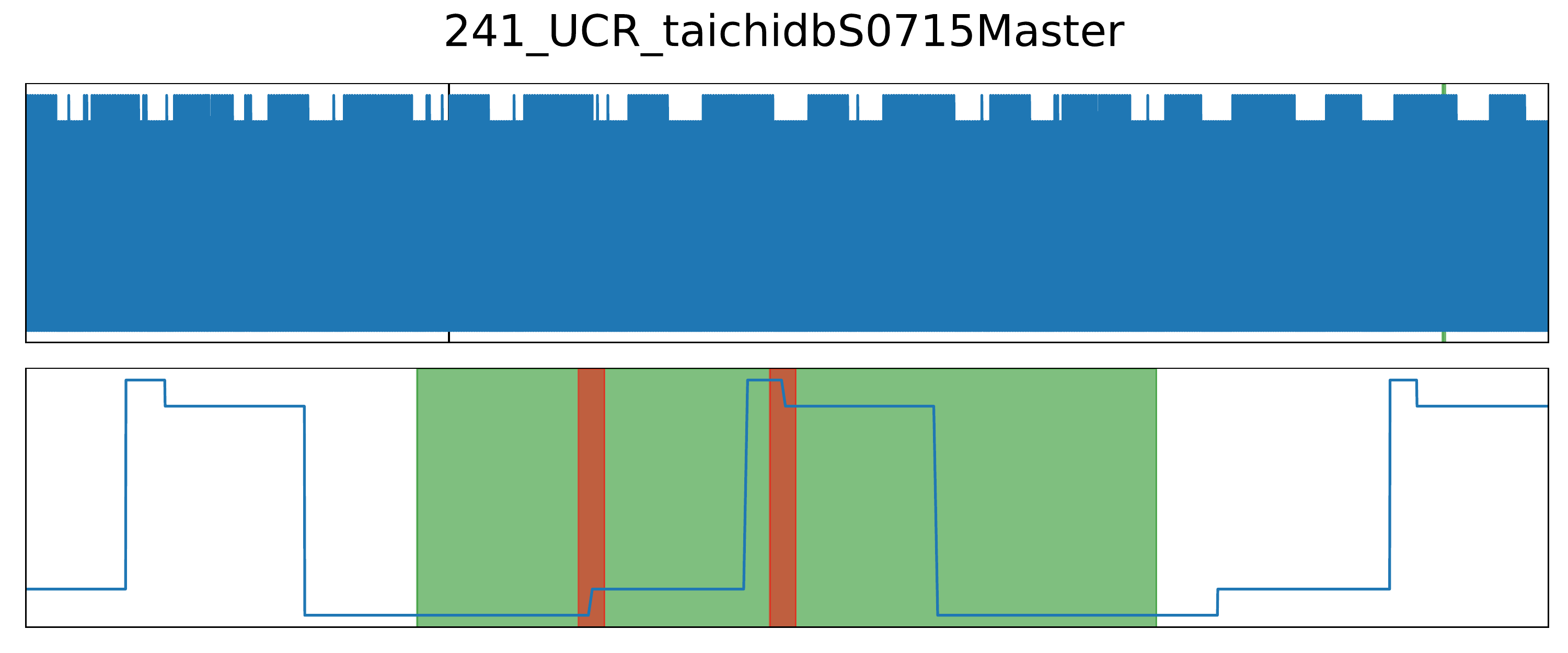}
        \caption{Overview (top) and detail (bottom) plot of the time series containing the ''smoothed increase'' anomaly. The ground truth anomaly is highlighted in green while the sections where the smoothing is visible are highlighted in red.}
        \label{fig.disc.smoothed_increase}
    \end{subfigure}
    \caption{Time Series containing  ''steep increase'' (a) and ''smoothed increase'' (b) anomalies.}
    \label{fig:increase_anomaly}
\end{figure}

The anomaly that was detected by all methods except RRCF is the ''steep increase'' anomaly shown in Figure~\ref{fig.disc.steep_increase}. 
This anomaly can be found in two time series of the UCR Anomaly Archive which represent the same data but was distorted in one case that is shown in Figure~\ref{fig.disc.steep_increase}. 
According to the UCR score, MDI and MERLIN detect this anomaly in every repetition of the experiment with the highest UCR score. 
As both methods are deterministic it is expected, that the results do not differ between multiple runs. 
AE finds this anomaly in 11/12 cases. 
GANF and TranAD detect the ''steep\_increase'' anomaly only in the undistorted version of the time series.
RRCF and TranAD detect the ''smoothed increase'' anomaly shown in Figure~\ref{fig.disc.smoothed_increase} where a normally steep increase was smoothed by increasing the number of different values in one cycle. 
For RRCF, this is also the only type this method can find. 
While RRCF has a UCR score below $0.05$ for 14/16 anomaly types, it detects the ''smoothed increase'' anomaly with a UCR score of $1.0$ and an F1 score of $0.86$. TranAD finds this anomaly in 5/6 cases.

For the remaining anomaly types, the results are more diverse. 
GANF, MERLIN and TranAD find the majority of the 23 ''outlier'' anomalies with GANF and TranAD performing better than MERLIN on this type. 
The 23 ''noise'' anomalies however are detected by AE, GANF and MDI with MDI finding every single one with the highest anomaly score. 

From a method point-of-view, MDI achieves UCR scores above or equal to 0.5 for the classes ''time warping'', ''steep increase'', ''sampling rate'', ''noise'', ''missing peak'', ''local peak'' and ''local drop''. 
For the ''noise'' type anomalies, the F1 score is above 0.5 as well. 
The tendency in the results for AE look similar to those of MDI but the scores for AE are mostly a few points lower, therefore AE has a UCR score above 0.5 only for ''noise'' and ''steep increase'' with the latter having an F1 Score of 0.92. 
MERLIN shows a UCR score above or equal to 0.5 for ''steep increase'', ''outlier'', ''missing drop'', ''local peak'', ''frequency change'' and ''amplitude change'' anomalies, making ''steep increase'' and ''missing peak'' the only classes where both methods have a UCR score above or equal 0.5. 
In terms of F1 Score, MERLIN scores above 0.5 for the classes ''noise'' and ''steep increase''. 
GANF is the best performing method regarding the anomaly types ''time warping'' and ''outlier'' with UCR scores of $0.75$ and $0.78$ and F1 scores of $0.78$ and $0.45$ respectively. Additionally, GANF detects at least half of the anomalies with type ''steep increase'' and ''noise''.
TranAD achieves a UCR score above or equal to 0.5 for the classes ''outlier'', ''smoothed increase'' and ''steep increase'' but for the latter, the corresponding low F1 Score indicates a high number of false positives. 
For the ''sampling\_rate'' anomalies it is vice versa, as the F1 Score is 0.61 here but the UCR score is 0.0. 
 
 We will discuss those differences in Section~\ref{sec:discussion} in more detail. 
 MDI and MERLIN together detect the anomalies of more than two third of the annotated anomaly types. 
 For the classes ''flat'', ''reversed'', ''time shift'' and ''unusual pattern'', no method achieved a UCR score above or equal to $0.5$.

\subsection{The influence of subsequence length on MDI and MERLIN}
\label{subsec:results:mdi_merlin}
The goal of this experiment was to examine the influence of the subsequence length on the results for MDI and MERLIN and to evaluate their ability to utilize additional information about the problem domain given with the range of subsequence lengths.
To make a fair comparison, we fixed the subsequence length range for MDI and MERLIN to $L_{min} = 75$ and $L_{max} = 125$ time steps in the results presented in Section~\ref{subsec:results:by_model} and ~\ref{subsec:results:by_anomaly_type}, regardless of the specific characteristics of the individual time series, such as cycle length or expected length of the anomaly.

We, therefore, compare the baseline results from Section~\ref{subsec:results:by_model} with two strategies for setting the subsequence range. 
For the "dynamic" strategy, we provided additional information by setting the range of subsequence lengths based on the length $L_{\Tilde{S}{a,b}} = b-a$ of the true anomaly $\Tilde{S}{a,b}$ to $L_{\Tilde{S}_{a,b}} \pm 25\%$.
For the "fixed" strategy, we chose a fixed length of 100 timesteps, thereby reducing the given information compared to the baseline.

\begin{figure}
\centering
\begin{subfigure}[b]{0.49\textwidth}
        \centering
        \includegraphics[width=\textwidth]{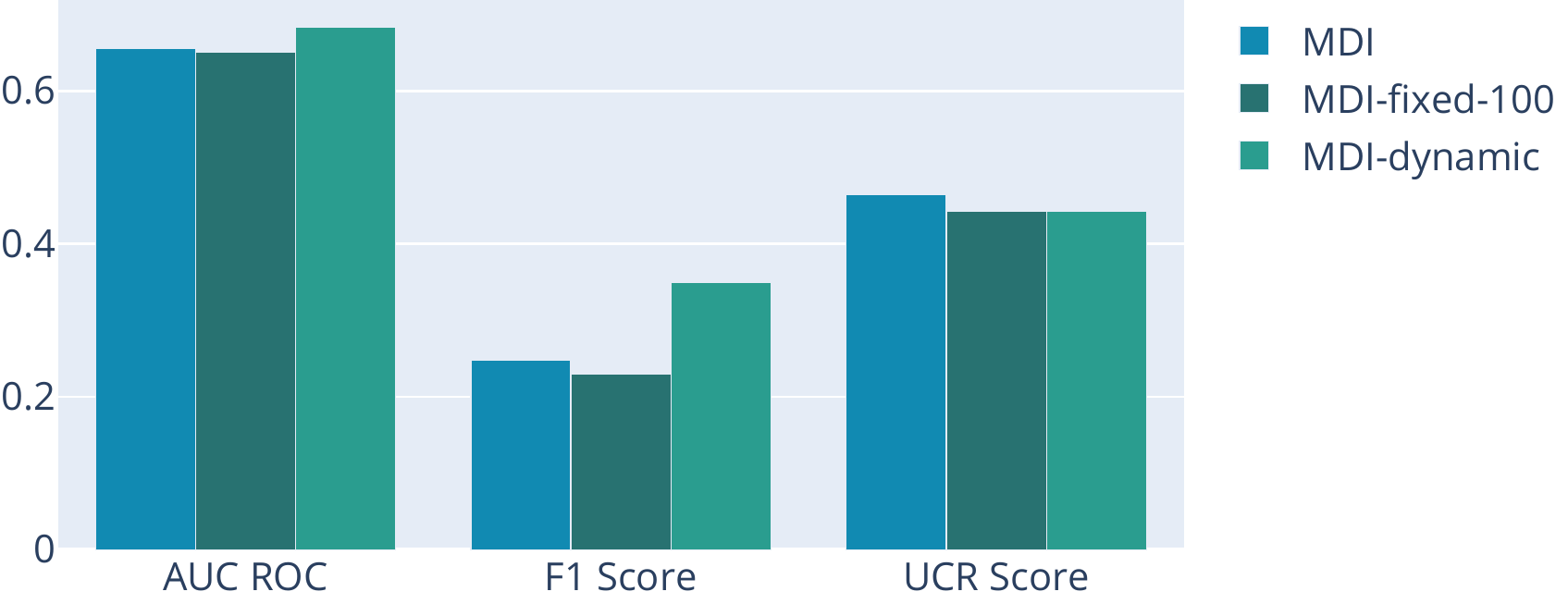}
        \caption{}
        \label{fig.results.mdi_metrics}
    \end{subfigure}
    \hfill
    \begin{subfigure}[b]{0.49\textwidth}
        \centering
        \includegraphics[width=\textwidth]{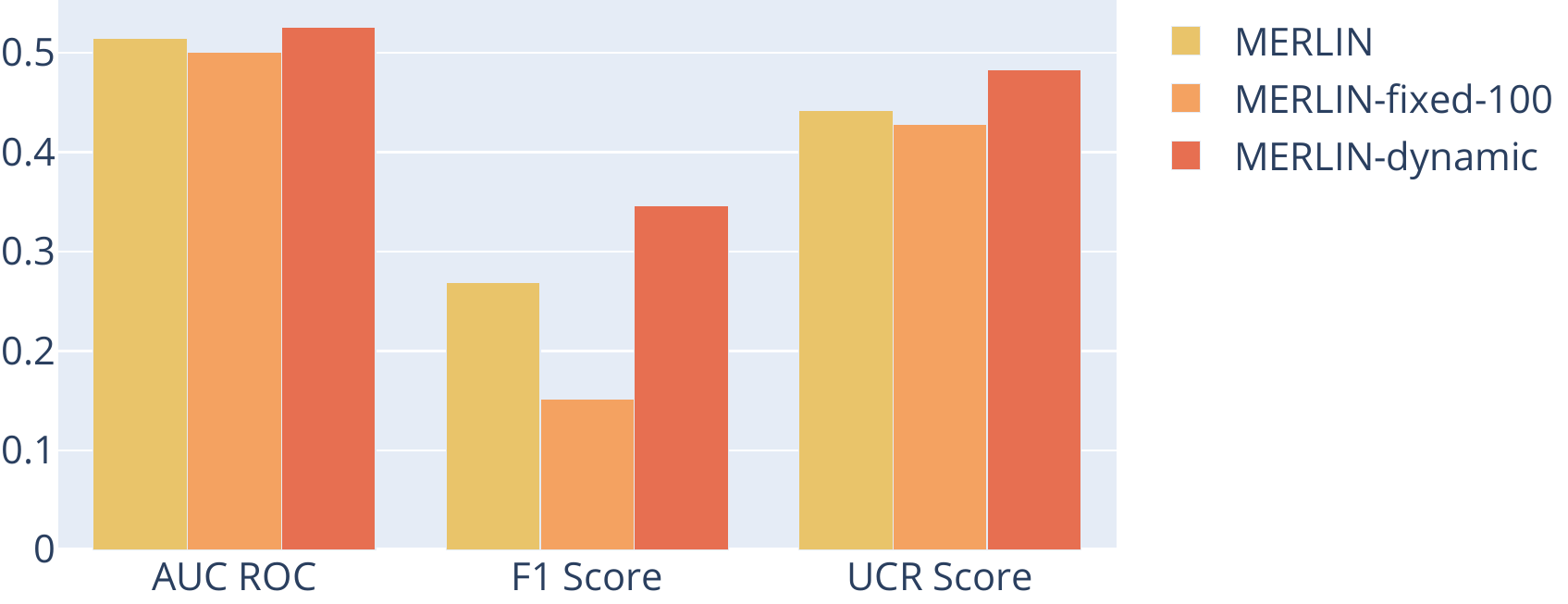}
        \caption{}
        \label{fig.results.merlin_metrics}
    \end{subfigure}
    \caption{Results for different subsequence lengths for MDI \textbf{(a)} and MERLIN \textbf{(b)}. The baseline results ''MDI'' and ''MERLIN'' are those from the main experiments using a subsequence length range of $L_{min} = 75$ and $L_{max} = 125$. The results labeled "MDI\_fixed\_100" and "MERLIN\_fixed\_100" were obtained using a fixed subsequence length of 100 time steps, while the results labeled "MDI\_dynamic" and "MERLIN\_dynamic" were obtained by individually setting $L_{min}$ and $L_{max}$ to 75\% and 125\% of the true anomaly length, respectively, for each time series.}
    \label{fig:mdi_merlin}
\end{figure}

The results for MDI, shown in Figure~\ref{fig.results.mdi_metrics}, demonstrate that fixing the subsequence length to 100 and reducing the given information leads to a decrease in the AUC ROC and F1 score.
In contrast, choosing the range for the subsequence length dynamically leads to an increase in the AUC ROC and F1 score.
The results for the UCR score do not reflect this trend; the highest UCR score is still achieved with the baseline configuration, but the difference between the fixed and dynamically chosen subsequence length is relatively small.

The results for MERLIN, displayed in Figure~\ref{fig.results.merlin_metrics}, show similar behavior, but the differences between the strategies are more pronounced.
For MERLIN, the positive effect of additional information is present across all three metrics.

\subsection{RRCF on sliding window statistics}
\label{subsec:results:rrcf}
\begin{figure}
\centering
    \begin{subfigure}[b]{0.45\textwidth}
        \centering
        \includegraphics[width=\textwidth]{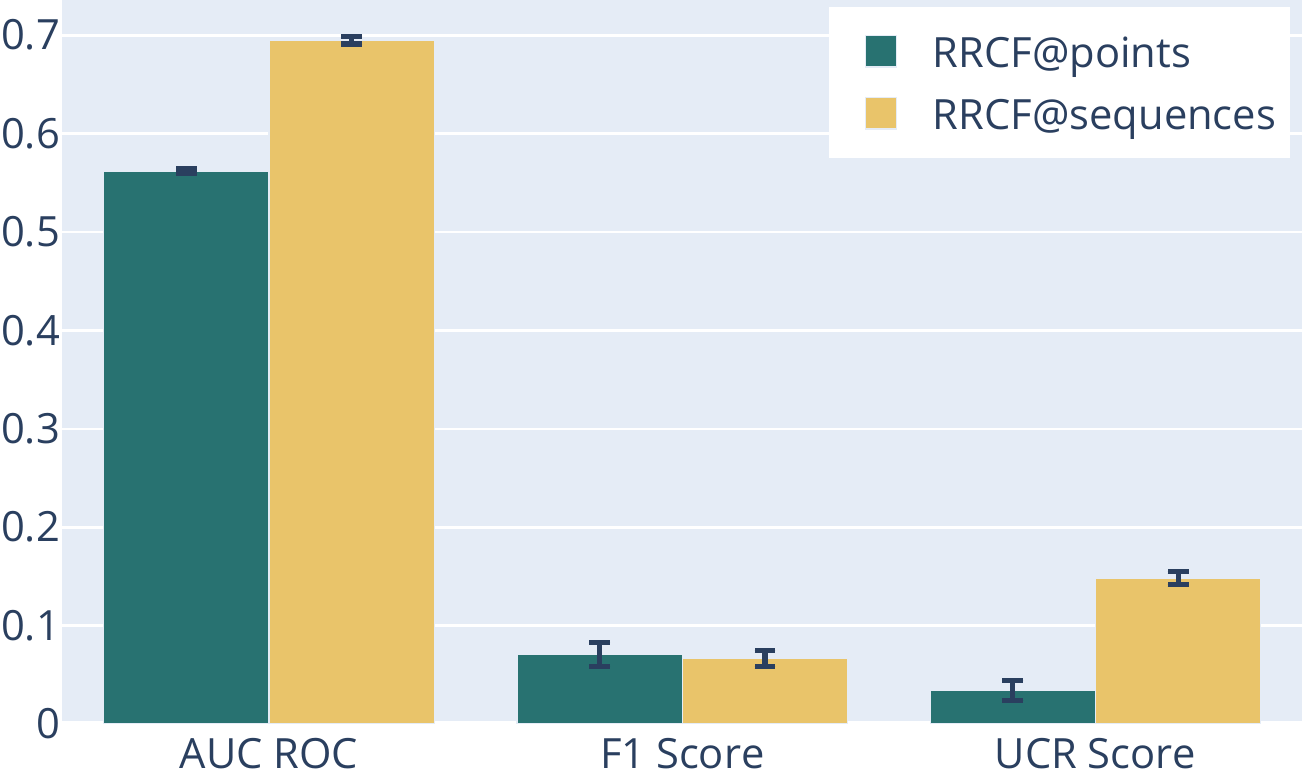}
        \caption{}
        \label{fig.results.rrcf_metrics}
    \end{subfigure}
    \hfill
    \begin{subfigure}[b]{0.53\textwidth}
        \centering
        \includegraphics[width=\textwidth]{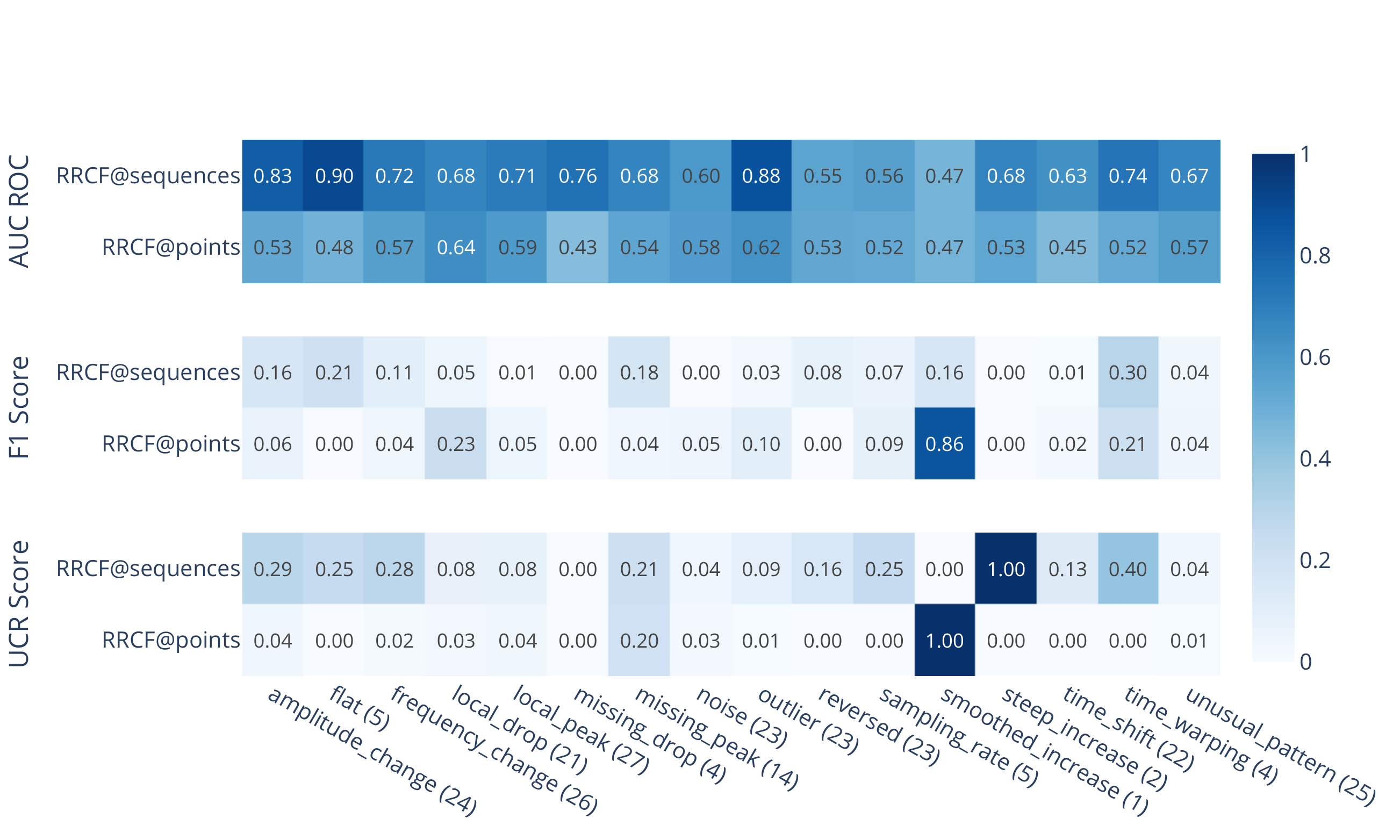}
        \caption{}
        \label{fig.results.metrics_by_anomaly_types_rrcf}
    \end{subfigure}
    \caption{Comparing AUC ROC, F1- and UCR score of the baseline RRCF using the raw time series as input (RRCF@points) to RRCF applied on sliding window statistics (RRCF@sequences). \textbf{(a)} Macro-averaged performance metrics for each method. The error bars indicate the standard deviation caused by random effects over six repetitions of the experiment. \textbf{(b)} Macro-averaged AUC ROC, F1 and UCR scores for the 16 annotated anomaly types over six repetitions of the experiment. Next to the anomaly type, the number of time series containing that type is shown in parenthesis.}
    \label{fig:results_rrcf_sequences}
\end{figure}

RRCF applied to point-wise features is tailored towards finding point anomalies due to its principle of isolating single points. 
In this experiment, we compare this baseline RRCF we used in the former experiments  (RRCF@points) to an alternative (RRCF@sequences) where we preprocess the time series by computing a vector consisting of the minimum, maximum, coefficient of variation and the first four moments (mean, variance, skewness, and kurtosis) of a sliding window. 
We choose a subsequence length of 100 and a stride of 50. 
We also tuned the hyper-parameters \textit{n\_trees} and \textit{tree\_size} as described in Section~\ref{subsec:setup}. 
The results are shown in Figure~\ref{fig:results_rrcf_sequences}. 
Using subsequence-wise features for RRCF increased the AUC ROC from $0.56$ to $0.7$ making this the best AUC ROC result among the analyzed methods. 
Also, the UCR score increased for RRCF@sequences by a factor of $5$ from $0.03$ to $0.15$. 
The F1 Score does not change substantially. 
While RRCF applied to point-wise features was the only method detecting the ''smoothed increase'' anomaly, this anomaly is not detected anymore. 
Instead, RRCF applied to subsequence-based features now detected the 'steep increase' anomalies like all other five methods. 
For all other anomaly types except ''midding drop'', the UCR score increase for RRCF@sequences. 
The highest increase is made for the ''time warping'' anomaly from $0.0$ to $0.4$.

\section{Discussion}
\label{sec:discussion}
Before discussing the results obtained for individual methods, it is necessary to explain how to interpret the various metrics and their combinations. 
The low macro-averaged scores across all methods shown in Table~\ref{tab:results_by_method} can have different causes. 

To understand those, we will build upon the discussion of the importance of jointly analyzing different metrics, given in Section~\ref{sec:methods:measures}, and focus on the various reasons for false positive or false negative results in the following.
For instance, a low F1 score may be due to an insufficient anomaly score, which prevents the detection of the true anomaly, or it may be due to a poor choice of threshold, leading to an increase in false positives.

\begin{figure}
\centering
    \begin{subfigure}[t]{0.47\textwidth}
        \centering
        \includegraphics[width=\textwidth]{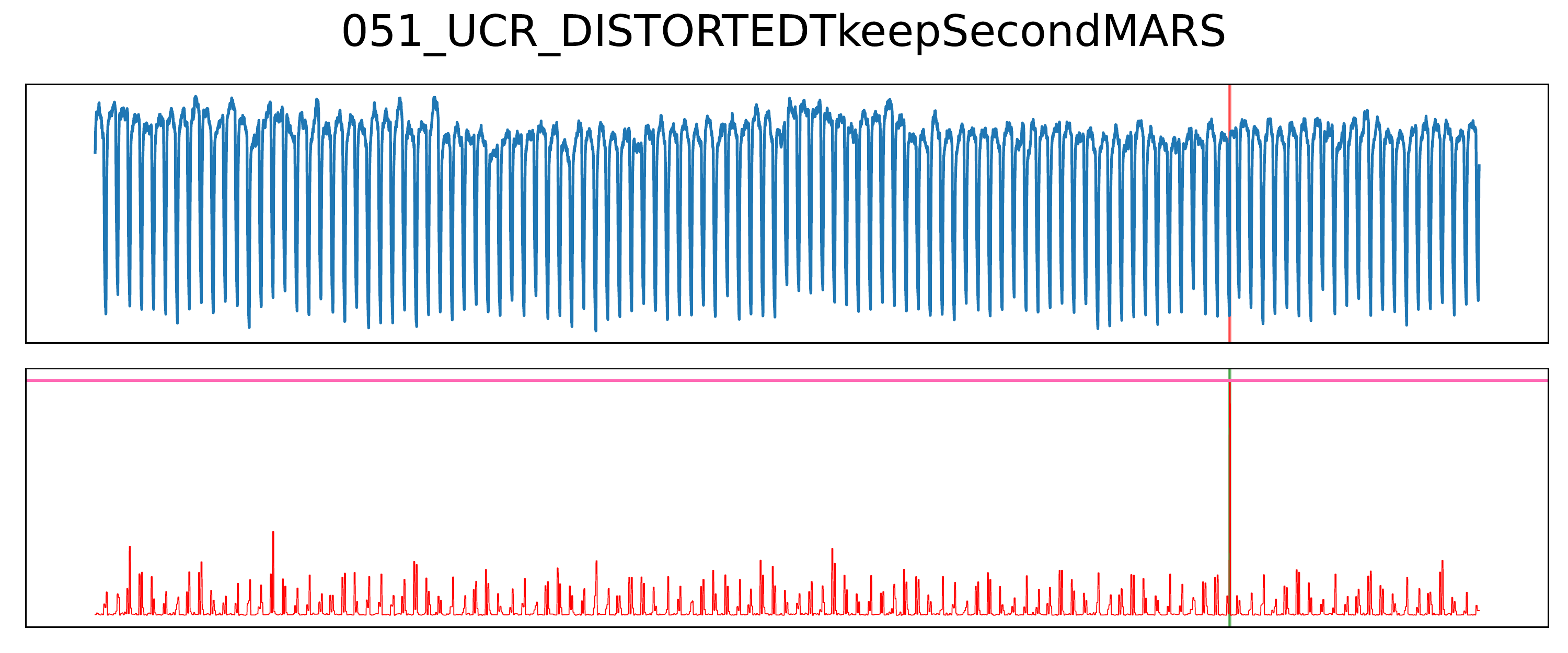}
        \caption{High AUC ROC, F1- and UCR score indicating correct detections without or with very few false positives.\\}
        \label{fig.disc.ae_all_scores_high}
    \end{subfigure}
    \hfill
    \begin{subfigure}[t]{0.47\textwidth}
        \centering
        \includegraphics[width=\textwidth]{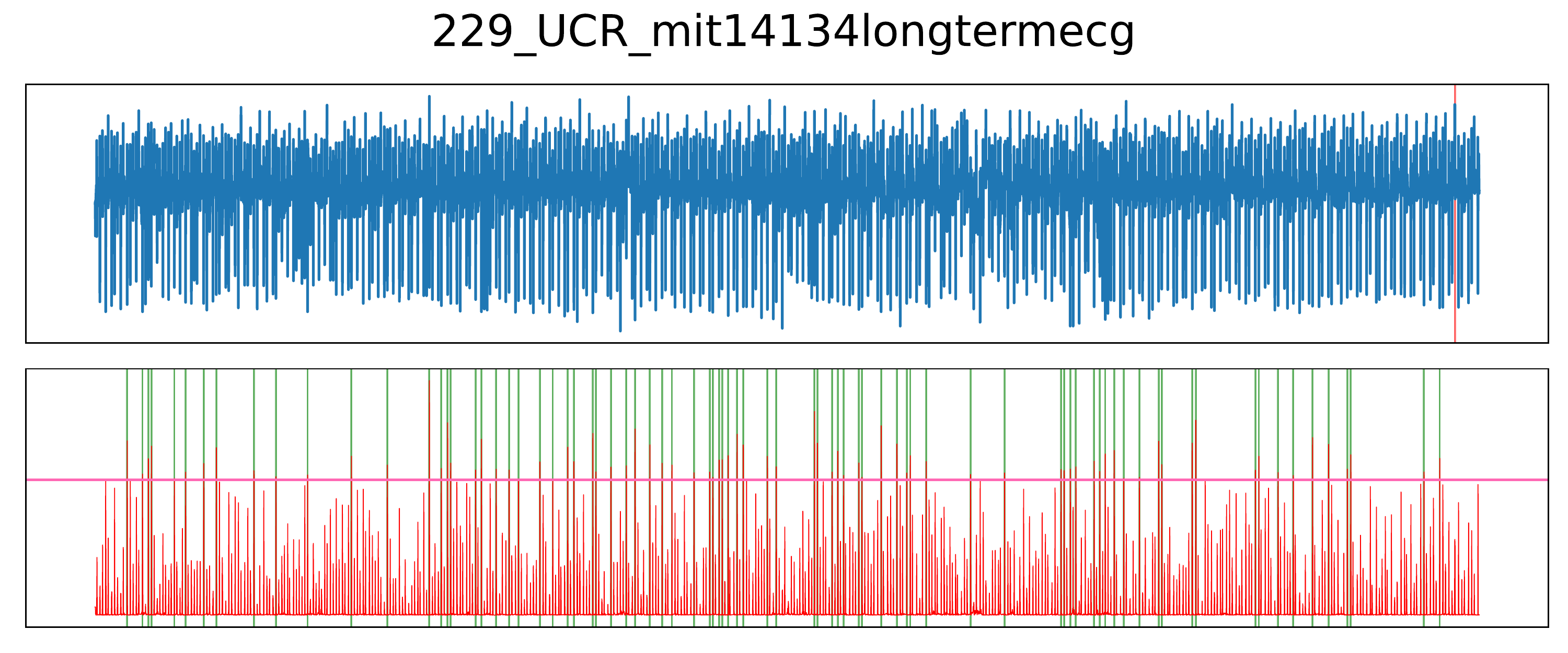}
        \caption{High AUC-ROC but low F1- and UCR score indicate an insufficient anomaly score and/or an insufficient threshold}
        \label{fig.disc.ae_high_auc_low_f1_low_ucr}
    \end{subfigure}
    \hfill
    \begin{subfigure}[t]{0.47\textwidth}
        \centering
        \includegraphics[width=\textwidth]{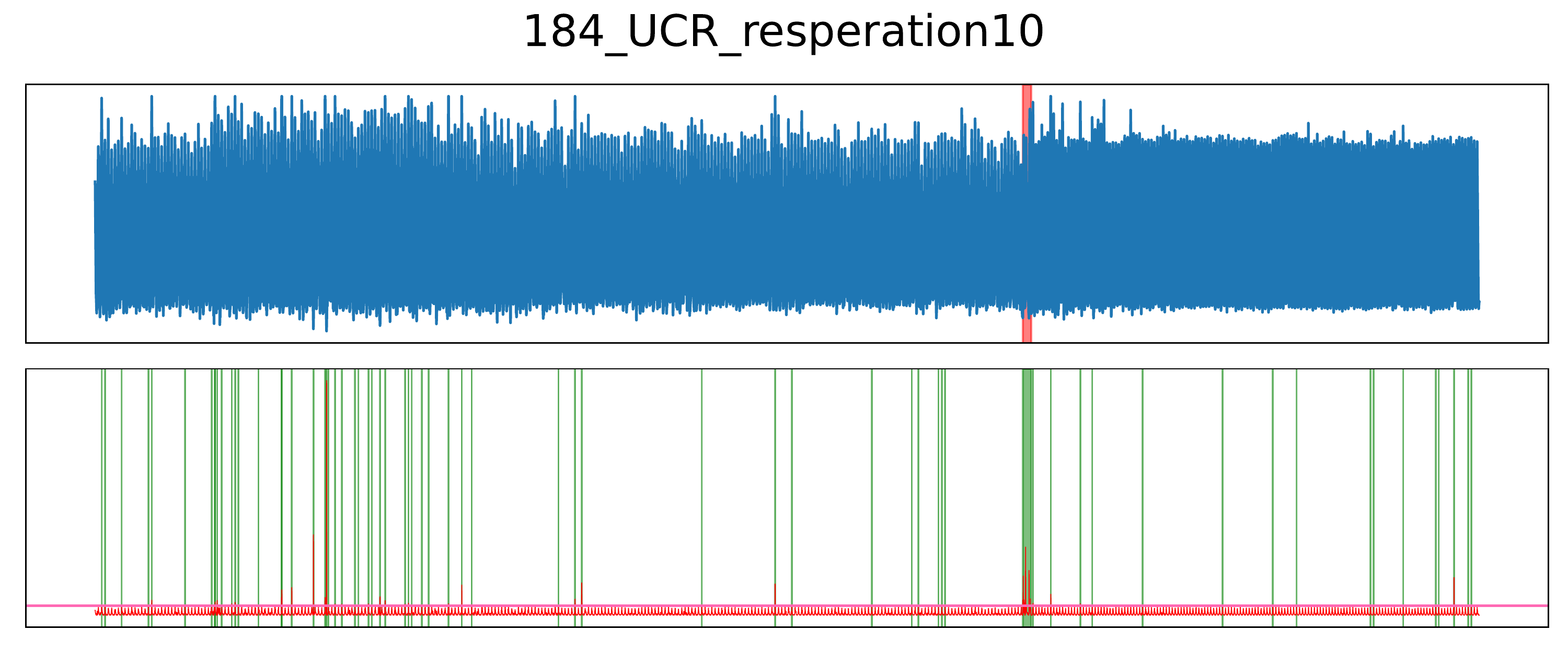}
        \caption{A high F1- but a low UCR score indicates the detection of the true anomaly but a false positive has a higher anomaly score.}
        \label{fig.disc_high_f1_low_ucr}
    \end{subfigure}
    \hfill
    \begin{subfigure}[t]{0.47\textwidth}
        \centering
        \includegraphics[width=\textwidth]{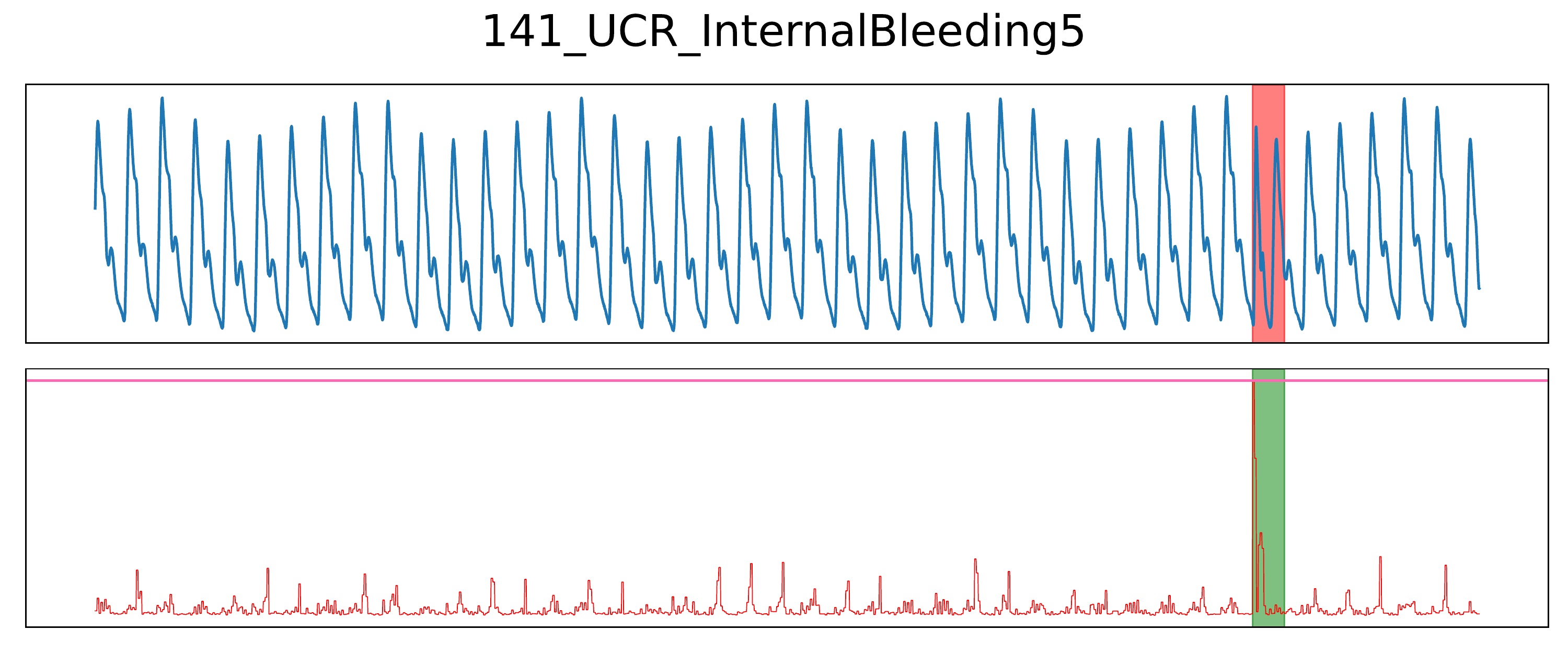}
        \caption{Low AUC ROC but high F1- and UCR score indicate a correct detection of the labeled anomaly without or with few false positives but the detected subsequence is shorter than the true anomaly.\\}
        \label{fig.disc.low_auc_high_f1_high_ucr}
    \end{subfigure}
    \hfill
    \begin{subfigure}[t]{0.47\textwidth}
        \centering
        \includegraphics[width=\textwidth]{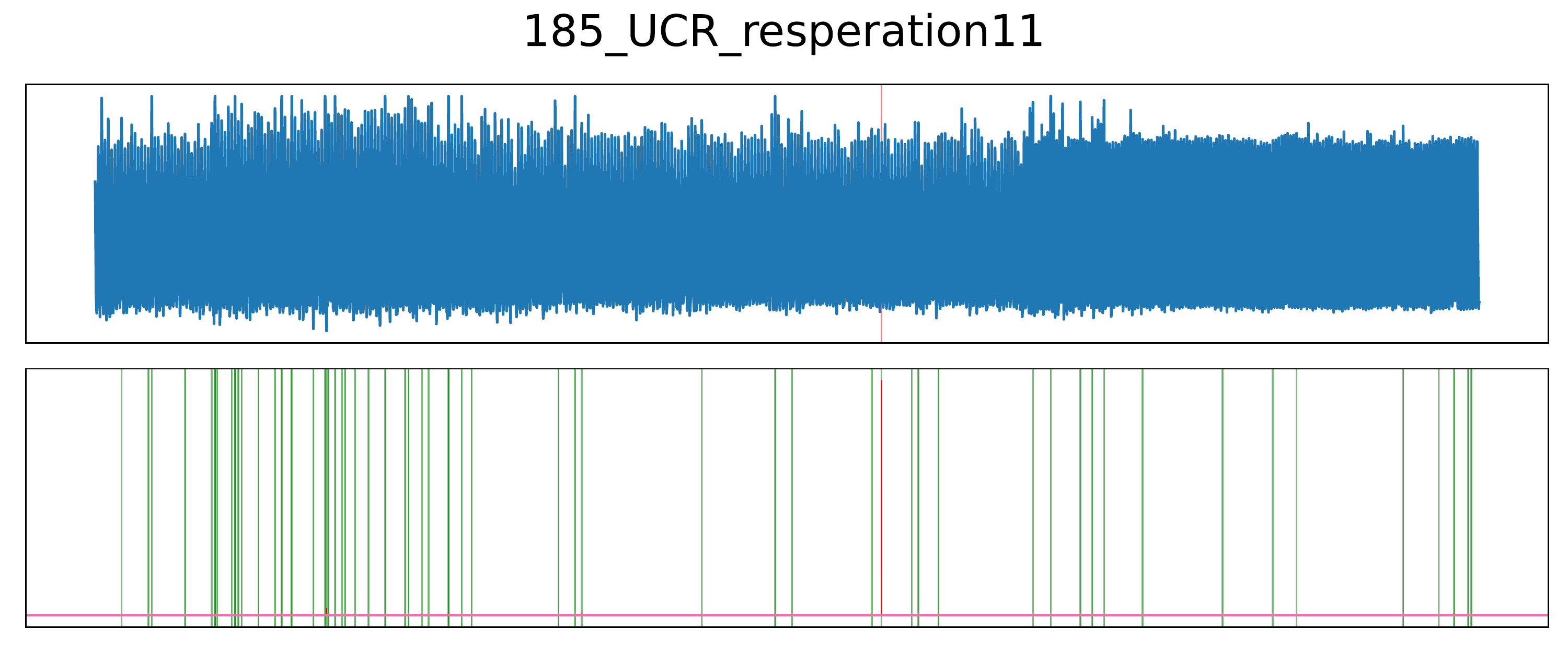}
        \caption{High UCR- but low F1 score indicates the correct detection of the true anomaly by the highest anomaly score but a bad threshold value leading to an increased number of false positives.}
        \label{fig.disc.mdi_low_f1_high_ucr}
    \end{subfigure}
    \hfill
    \begin{subfigure}[t]{0.47\textwidth}
        \centering
        \includegraphics[width=\textwidth]{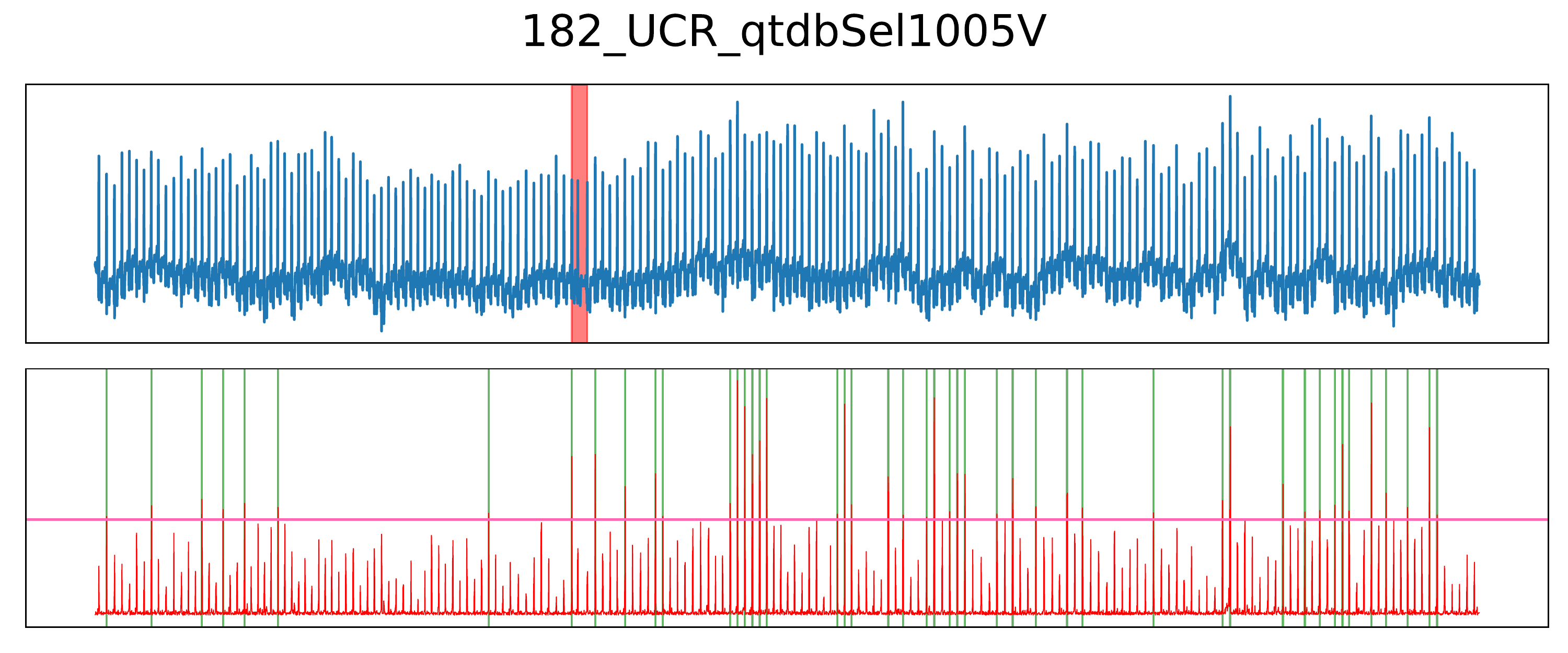}
        \caption{Low AUC ROC, F1- and UCR scores indicate an insufficient anomaly score not detecting the true anomaly.}

        \label{fig.disc.ae_all_low}
    \end{subfigure}
    
    \caption{Interpretation of results shown on examples of the autoencoder model. The top figure represents the time series with the true anomaly marked with red while the bottom figure shows the anomaly score with the pink horizontal line being the determined threshold and the predicted anomalies highlighted in green.}
    \label{fig:ae_high_and_low_scores}
\end{figure}

Figure~\ref{fig:ae_high_and_low_scores} illustrates this using different results for the autoencoder model. 
A high F1 score and a UCR score of 1 at the same time indicate the successful detection of the true anomaly without any, or with very few, false positive results, depending on the value of the F1 score, as shown in Figures~\ref{fig.disc.ae_all_scores_high} and \ref{fig.disc.low_auc_high_f1_high_ucr}.
On the other hand, a low F1 score and a UCR score of 0, as shown in Figures~\ref{fig.disc.ae_high_auc_low_f1_low_ucr} and \ref{fig.disc.ae_all_low}, indicate that the anomaly was not detected due to an insufficient anomaly score. 
In this case, a high AUC ROC value may indicate that the anomaly could have been detected with a low anomaly score, but the threshold was set too high, resulting in the subsequence not being classified as anomalous. 
Figures~\ref{fig.disc.ae_high_auc_low_f1_low_ucr} and \ref{fig.disc.ae_all_low} also demonstrate that AUC ROC is generally not a suitable measure to assess the quality of results in highly unbalanced problems in a meaningful way. 
In both cases, the anomaly score is not suitable for detecting the true anomaly.

\begin{figure}
\centering
    \begin{subfigure}[t]{0.47\textwidth}
        \centering
        \includegraphics[width=\textwidth]{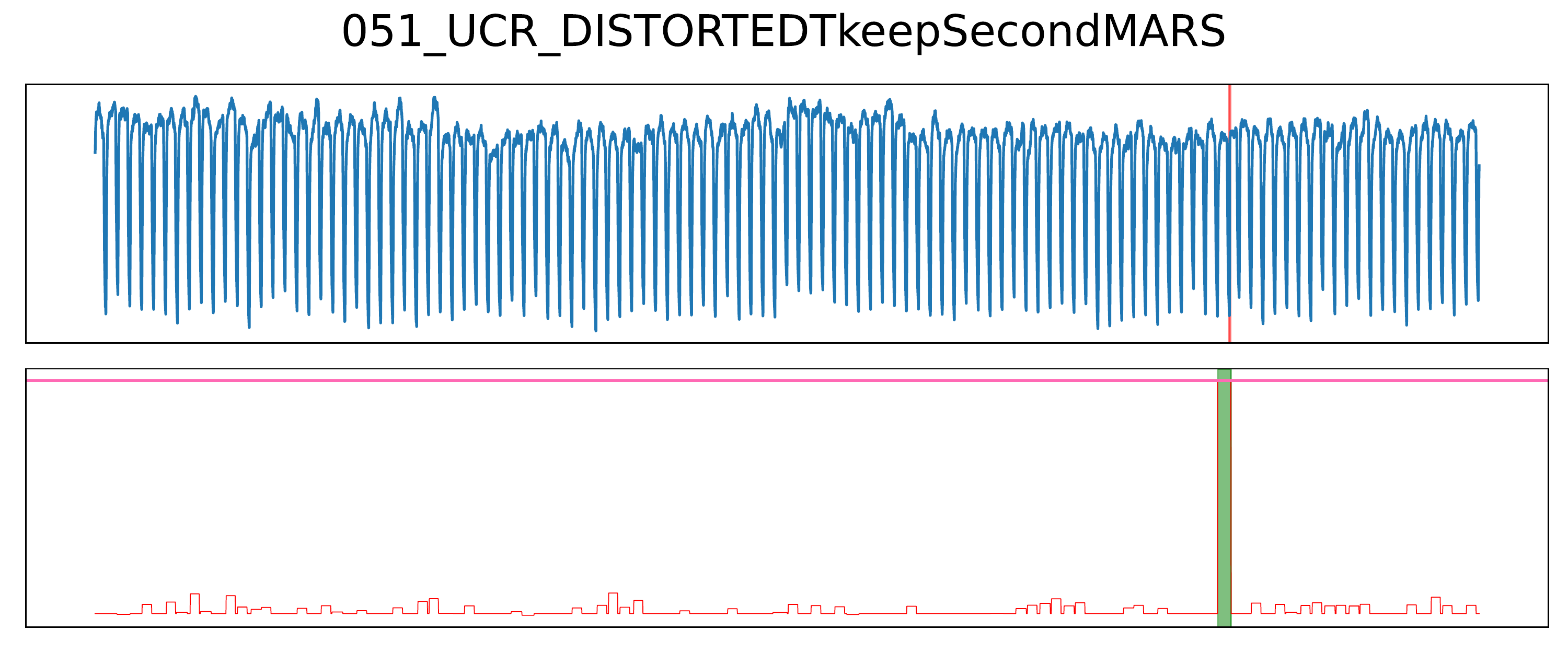}
        \caption{Low F1 score caused by false positive results due to the length of the correct detected anomaly.}
        \label{fig.disc.steep_increase_mdi}
    \end{subfigure}
    \hfill
    \begin{subfigure}[t]{0.47\textwidth}
        \centering
        \includegraphics[width=\textwidth]{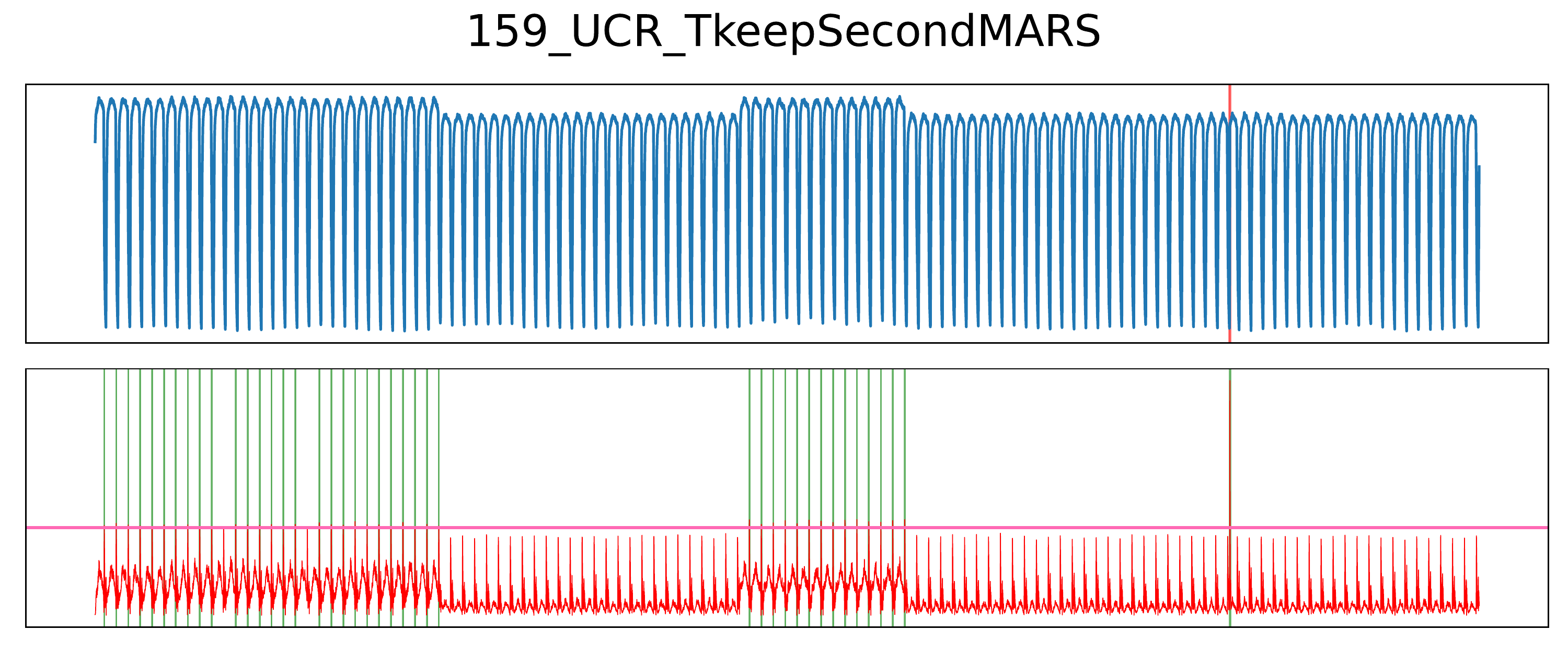}
        \caption{Low F1 score caused by false positive results due to a poor threshold.}
        \label{fig.disc.steep_increase_tranad}
    \end{subfigure}
    \caption{Two different reasons for low F1 scores - (a) shows a result from MDI, (b) shows a result from TranAD.}
    \label{fig:steep_increase_mdi_}
\end{figure}

\begin{figure}
\centering
    \begin{subfigure}[t]{0.47\textwidth}
        \centering
        \includegraphics[width=\textwidth]{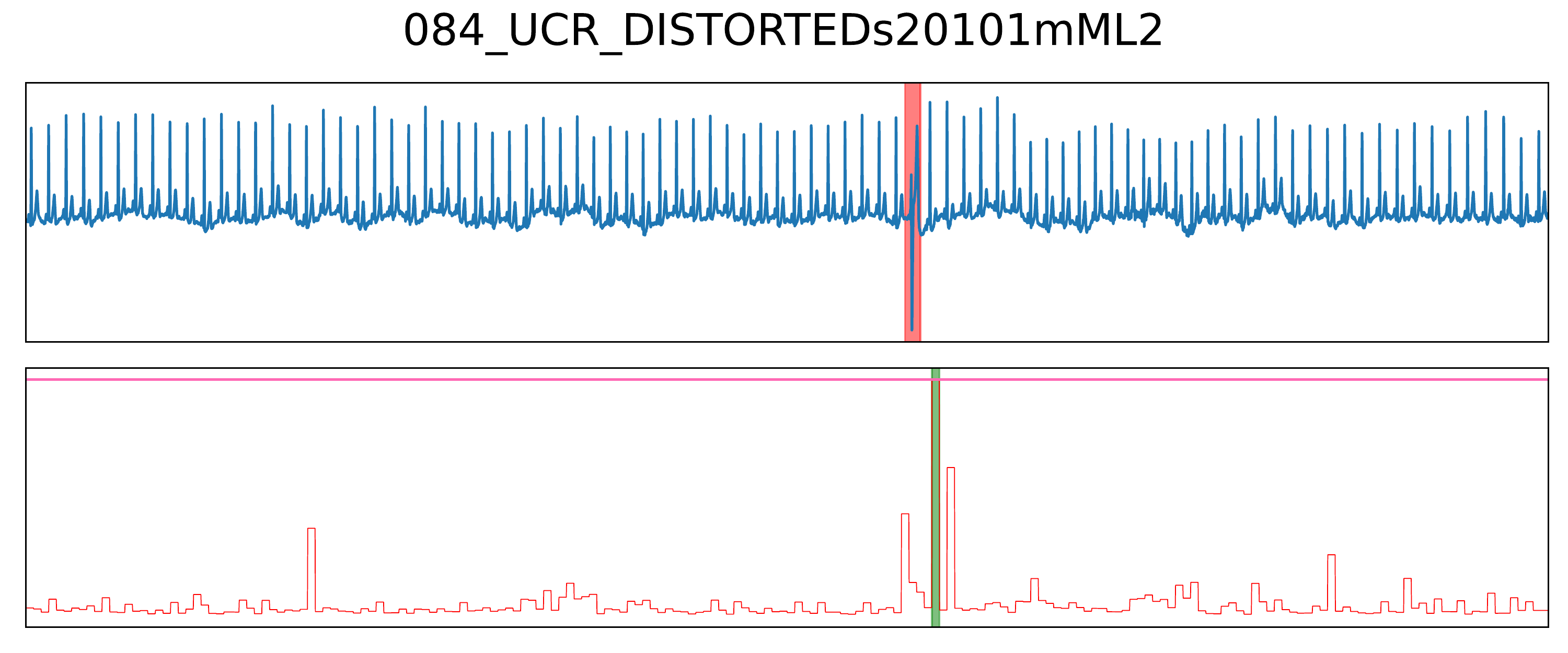}
        \caption{The anomaly is detected slightly after the ground truth label but within the 100 time steps tolerance for short anomalies in the UCR score.}
        \label{fig.disc.rrcf_sequences_too_late}
    \end{subfigure}
    \hfill
    \begin{subfigure}[t]{0.47\textwidth}
        \centering
        \includegraphics[width=\textwidth]{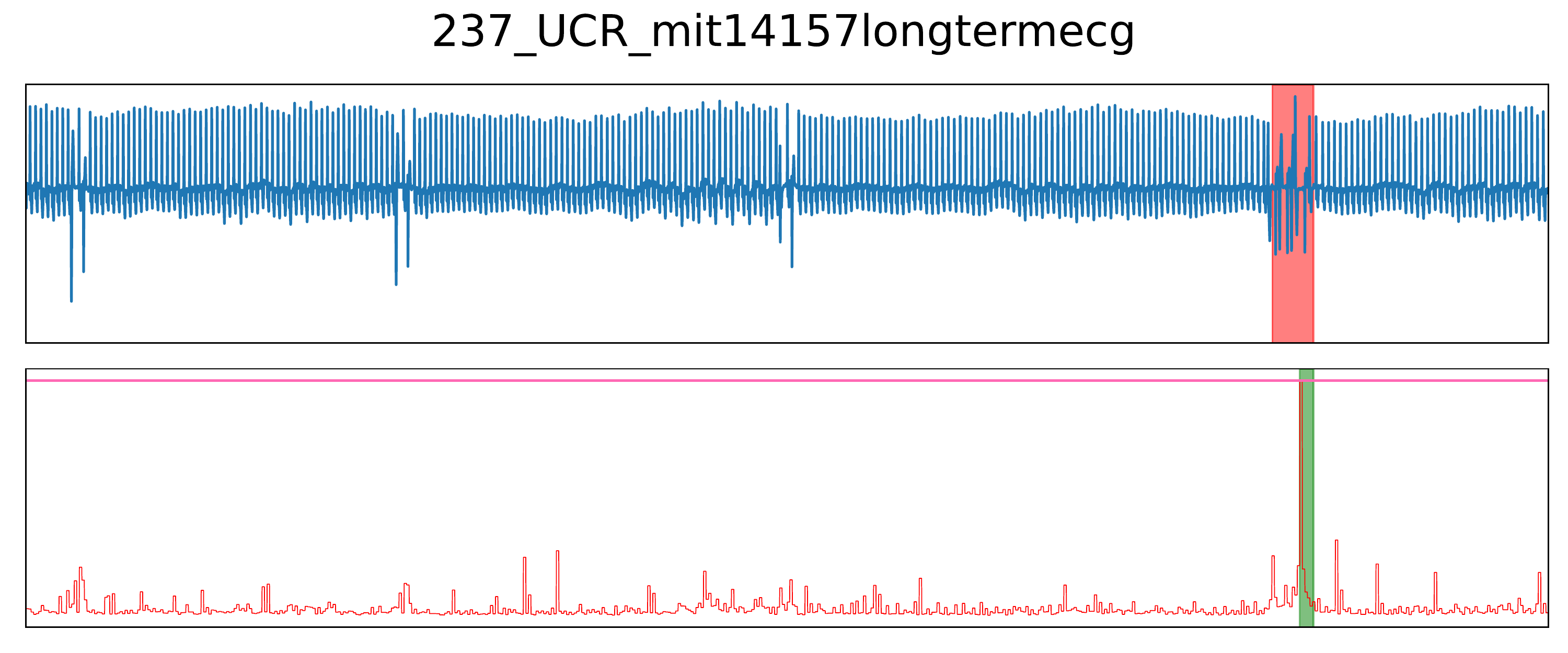}
        \caption{The detected anomaly is shorter than the ground truth label causing false negatives in the computation of the F1 score.}
        \label{fig.disc.rrcf_sequences_too_short}
    \end{subfigure}
    \caption{Two different reasons for low F1 scores. Both results are from RRCF.}
    \label{fig:disc:rrcf_sequences}
\end{figure}

A high F1 score but a UCR score of 0, as shown in Figure~\ref{fig.disc_high_f1_low_ucr}, indicates that the true anomaly was detected, but a false positive result has a higher anomaly score. 
A UCR score of 1 but a low F1 score signifies the correct detection of the true anomaly with the highest anomaly score but false positive or false negative results lead to a low F1 score. Figure~\ref{fig:steep_increase_mdi_} and Figure~\ref{fig:disc:rrcf_sequences} illustrate the different reasons for this situation, which can be caused by a poor threshold value, as shown in Figure~\ref{fig.disc.steep_increase_tranad}, or the detected anomaly's subsequence length being much longer as in Figure~\ref{fig.disc.steep_increase_mdi} or shorter than the ground truth label as in Figure~\ref{fig.disc.rrcf_sequences_too_short} leads to increased false positive or false negative results, respectively.
A fourth case with this result occurs from the detection of a short anomaly within the 100 time steps tolerance, which is considered in the definition of the UCR score in Equation~\ref{eq:ucrscore}. 
This is shown in Figure~\ref{fig.disc.rrcf_sequences_too_late}.

The best results in terms of F1 score and UCR score are obtained by MDI and MERLIN, with MERLIN having a slightly higher F1 score and MDI scoring slightly higher in terms of UCR score. 
The differences between these two methods are around 0.02. 
This difference in F1 score is likely due to the different methods used to choose the threshold. 
Both methods return a score only for the detected anomalous sequences, but MDI either requires the number of anomalies to be returned or may return a score for up to every subsequence. 
In order to not give MDI an advantage over its competitors, the latter option was chosen and a threshold was determined using the POT method instead of the minimum anomaly score, as was done for MERLIN. 
This makes MDI more prone to detecting false positives in the described setup compared to MERLIN, which only returns the subsequences that have been detected as anomalous. 
RRCF performs poorly in terms of F1 score and UCR score, which may be due to its mechanism for isolating single points and its focus on point anomalies.

Among the deep learning methods, GANF demonstrates the best performance.
GANF achieves the highest scores for all three metrics and detects the largest variety of anomaly types.
The results for AE and TranAD are mixed.
AE has a higher AUC ROC and UCR score, but only detects two different anomaly types with a UCR score above 0.5. 
TranAD, in contrast, has a slightly higher F1 score when compared to AE and detects three different anomaly types.

Although the low results for RRCF, the classical machine learning methods show superior performance compared to the deep learning methods when aggregating the results by method class as shown in Figure~\ref{fig.results.by_modelclass}. 
This difference is particularly notable in the F1 score.
One assumption might be that these methods perform better in unsupervised settings that do not require a training phase.
However, this is contradicted by the RRCF results.

In terms of runtime, MERLIN has the longest average processing time of 291 seconds per time series.
This runtime is mainly determined by the discord discovery algorithm, which is called for every subsequence length $L_i \in \lbrace{L_{min}, ..., L_{max}\rbrace}$.
The complexity of this algorithm is quadratic with respect to the size of the set of potential discords, which is determined in the candidate selection phase.
However, for small candidate sets produced by a "good" choice for the parameter $r$ \cite{Nakamura2020}, the complexity becomes effectively linear.
As MERLIN starts with the highest possible value for $r$ and decreases it, it is unlikely to encounter a case where a small $r$ value causes the candidate subset to become too large.
On the other hand, MDI uses a subsequence proposal technique based on Hotelling's $T^2$ method \cite{MacGregor1994}, which selects interesting subsequences based on point anomaly scores rather than performing full scans over the data.
In addition to these differences in candidate subsequence selection, the specific implementations of the algorithms also have a significant impact on their runtime.
While we implemented MERLIN purely in Python, MDI is implemented in C++ with a Python interface.

All methods except RRCF were able to detect the "steep increase" anomaly shown in Figure~\ref{fig.disc.steep_increase} with varying UCR scores ranging from $0.5$ for GANF and TranAD to $1.0$ for MDI and MERLIN.
The low F1-Score for MDI indicates a high number of false positive results, which in this case depends on the length of the detected subsequence as shown in Figure~\ref{fig.disc.steep_increase_mdi}.
In contrast, the low F1 score for TranAD is caused by a poor threshold, leading to an increased number of false positive results as shown in Figure~\ref{fig.disc.steep_increase_tranad}.

RRCF is unique in detecting the ''smoothed increase'' anomaly shown in Figure~\ref{fig.disc.smoothed_increase} but not the ''steep increase'' anomaly like the other five methods. 
This behavior can be explained by the working principle of RRCF to isolate single points. 
The values in the smoothed subsequence occur only once in the time series and can therefore be isolated from all other values. 
That RRCF does not find the outlier anomalies seems contradicting but is due to the time series containing other extreme values, e.g. with an inverted sign, covering the true anomaly in the anomaly score.

The comparison of different strategies for choosing the range of subsequence lengths for MDI and MERLIN presented in Section~\ref{subsec:results:mdi_merlin} reveals that both methods can utilize additional information about the anomalies, with a stronger effect for MERLIN.
Providing additional information in terms of the subsequence length of the true anomaly increased the F1 score for MDI but decreased the UCR score by $0.02$ which indicates that MDI utilized the additional information to reduce false positive results. 
For MERLIN the F1 score and UCR score increased, indicating that the information on the true anomaly length helped MERLIN to identify anomalies it missed before.

In the final experiment, we used subsequence-based statistics instead of point-wise features for RRCF, which increased the UCR score and the AUC ROC. 
However, the macro-averaged F1 score slightly decreased due to the inability to detect the "smoothed increase" anomaly.
Instead, RRCF@sequences was able to detect the "steep increase" anomaly like the other five methods, indicating that this anomaly can only be detected on the subsequence level.
The low F1 scores for RRCF@sequences on those time series with a UCR score of 1 are mostly caused by the anomaly being detected slightly before or after the ground truth label but within the 100 time steps tolerance for short anomalies, or by the true anomaly being much shorter or longer than the subsequence length used for RRCF@sequences. 
Figure~\ref{fig:disc:rrcf_sequences} illustrates these two cases.

We conclude this section by summarizing the strengths and weaknesses of the methods analyzed in this study.
MDI and MERLIN have the notable advantage of not requiring any hyperparameter tuning. 
The only parameters that need to be set are the minimal and maximal subsequence lengths, which practitioners select based on the specific application or domain.
Despite the arbitrary choice of $L_{min} = 75$ and $L_{max} = 125$ time steps, MDI and MERLIN still outperform all other methods in this study. 
Additionally, these methods detect a wide range of anomaly types. 
However, a disadvantage of MDI and MERLIN is that they are not immediately applicable in an online setting.
Although discord discovery can be performed online using a different algorithm like DAMP \cite{Lu2022}, MERLIN cannot be directly applied to data streams. Similarly, while it may be possible to adapt MDI to consider only subsequences up to a given timestamp when estimating the density of $\Omega(S)$, the current version of MDI does not support this.

The isolation forest approach used in RRCF is intuitive and can be applied to data streams, which are advantages of RRCF. 
However, RRCF shows poor results in this study and may be more suitable for applications where outliers have distinct values from normal data. 
All three classical methods have the advantage of being easily interpretable.

GANF is the best-performing deep learning-based method in this study, which suggests that density estimation-based methods are effective in detecting anomalous sequences. 
Additionally, GANF is capable of being applied online once trained and has the potential to learn the dependency graph of multiple time series, which, although not analyzed in this study, could be beneficial in specific applications. 
A major disadvantage of GANF is the need to select values for numerous hyperparameters.
In our experiments, we used Bayesian Optimization to determine suitable values for the three most important hyperparameters (latent space dimension, learning rate and number of blocks), as identified by \cite{Dai2021}, using 10\% of the time series in the UCR Anomaly Archive. 
We used the default values from \cite{Dai2021} for the remaining eight hyperparameters, as they had not been tuned in that study either.
Using default values from \cite{Dai2021} for the three tuned hyperparameters leads to a decrease in the F1 score of 2\%-5\% and a drop in the UCR score of up to 19\%. 
However, averaging the two sets of hyperparameter values used in \cite{Dai2021} increases the UCR score by approximately 6\%. 
These better hyperparameters were not identified during the hyperparameter search, which highlights the general disadvantage of methods with a high number of hyperparameters.

The results for AE and TranAD are inconclusive but generally worse when compared to GANF. 
However, they are not as poor as the results for RRCF. 
Additionally, these two methods also have the disadvantage of having various hyperparameters that need to be set. 
When using the default parameters from \cite{Tuli2022} for TranAD, the results for F1- and UCR score decrease by about 4\% - 2\%, depending on which set was used. 
The values mentioned in the paper differ from those used in the repository. 
For AE, we do not have a set of default parameters, but we observed comparable or slightly worse results when choosing an arbitrary set of parameters. 
Like GANF, both methods have the advantage of being able to be applied to data streams after being trained.

\section{Conclusions}
\label{sec:conclusion}
In this study, we compared six anomaly detection methods, three of which were classical machine learning methods and three of which were based on deep learning.
We conducted extensive experiments on the UCR Anomaly Archive benchmark dataset, which we annotated with the types of anomalies present. 
We compared the methods on both, a dataset level and an anomaly-type level, to address two main questions: 
Does the potential superior performance of deep learning methods justify the sacrifice of the intrinsic interpretability of classical methods? 
And what are the similarities and differences between the analyzed methods in detecting different anomaly types? 
Our experiments showed that the classical machine learning methods MDI and MERLIN outperform the deep learning methods. 
The third classical method, RRCF, was unable to detect a substantial number of anomalies but improved when using sequence-based statistical features instead of raw data points. 
Among the deep learning methods, the Autoencoder model detected the most anomalies and was also the simplest model in this group.

While we present our experimental results in this work, a deeper theoretical analysis of the reasons and mechanisms behind these results is left for future research.
Regarding the second question about the similarities and differences in detecting certain anomaly types, we found that all subsequence-based methods detect the "steep increase" anomaly but not the "smoothed increase," while the opposite is true for the method that uses point-wise features. 
However, these classes are too small to produce a significant result. 
Although MDI and MERLIN had the best results in this comparison, they detected a diverse range of anomaly types. 
Together, they detected most of the anomalies, i.e., they detected 11 out of 16 anomaly types. 
However, the anomaly types "unusual pattern," "time shift," "reversed," and "flat" could not be reliably detected by any of the analyzed models.
A more theoretical analysis of these results will be conducted in a subsequent study.








\clearpage

\appendix
\section{Supplemental Materials}

\subsection{Abbreviations}
The following abbreviations are used in this manuscript:\\
\begin{tabular}{@{}ll}
    ABP & Arterial Blood Pressure \\
    AE & Autoencoder \\
    ECG & Electrocardiogram \\
    EPG & Electrical Penetration Graph \\
    EVT & Extreme Value Theory \\
    GANF & Graph Augmented Normalizing Flows \\
    ICP & Intracranial Pressure \\
    MDI & Maximally Divergent Intervals \\
    POT & Peak Over Threshold \\
    RRCF & Robust Random Cut Forest \\
    TranAD &  Transformer Network for Anomaly Detection
\end{tabular}

\subsection{}
\label{appendix:anomaly_types}
\begin{longtable}{|m{2.5cm}|m{3.5cm}|m{6cm}|}
\hline
\textbf{Anomaly Type} & \textbf{Description} & \textbf{Example} \\ 
\hline
\endhead
\hline
\endfoot
\endlastfoot
Amplitude Change  & Amplitude of the signal increased or decreased within a section.                                          & \includegraphics[width=\linewidth]{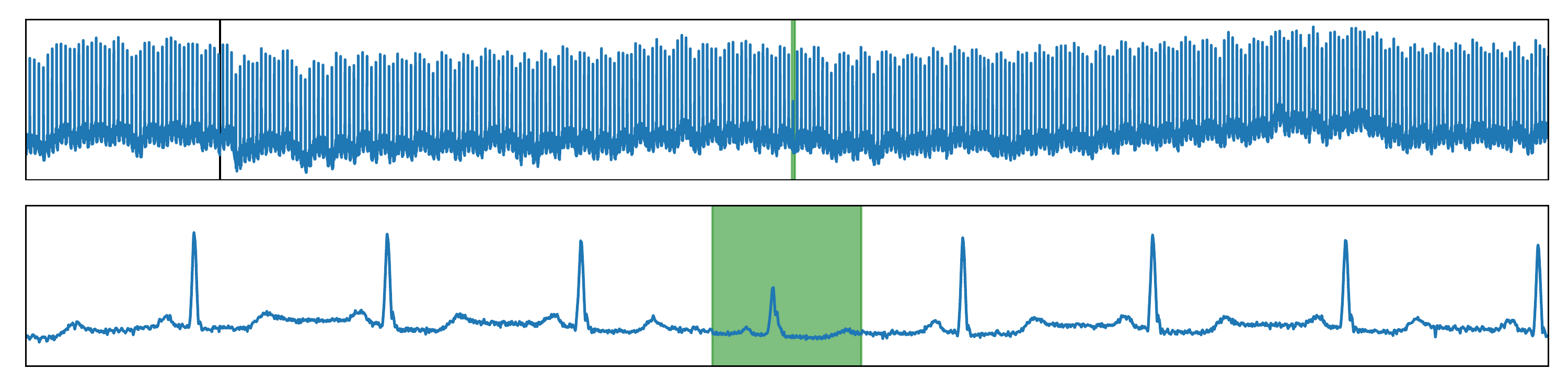}           \\ \hline
 Flat              & Flat section was added.                                                                                   & \includegraphics[width=\linewidth]{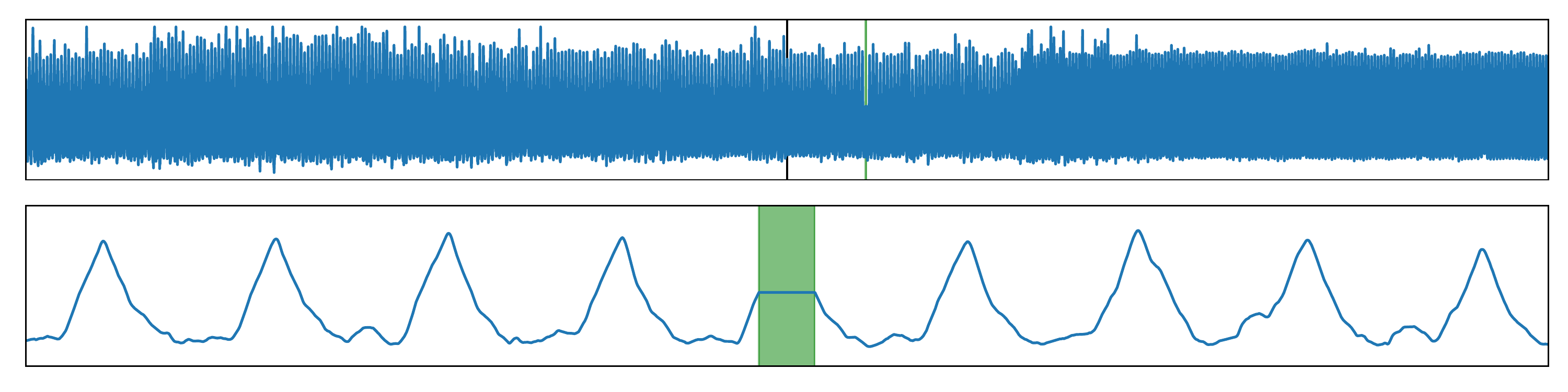}              \\ \hline
 Frequency Change  & The cycle length was modified within a section.                                                           & \includegraphics[width=\linewidth]{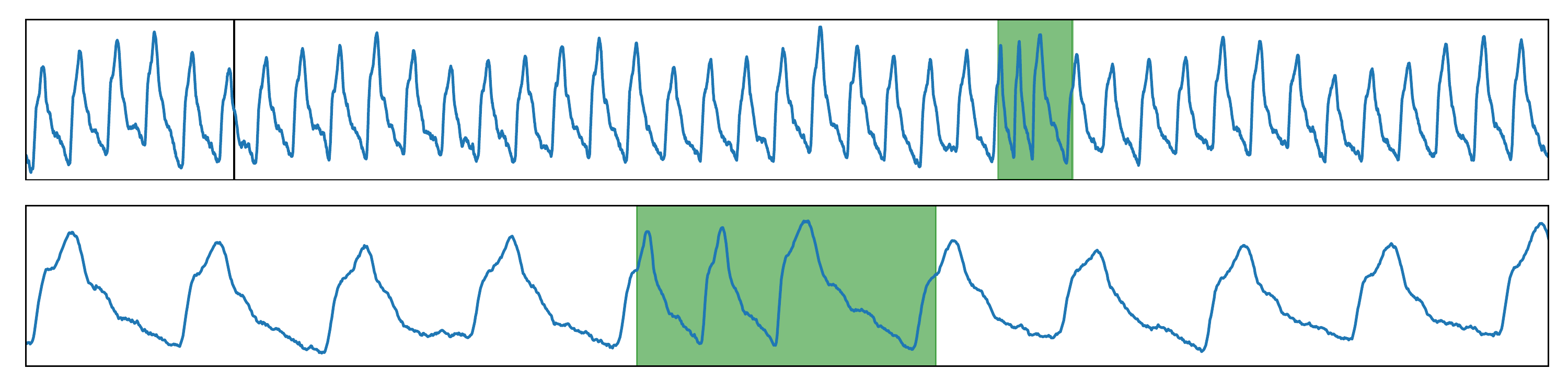}      \\ \hline
 Local Drop        & A drop was added, which is shallower than the minimal value of the time series.                           & \includegraphics[width=\linewidth]{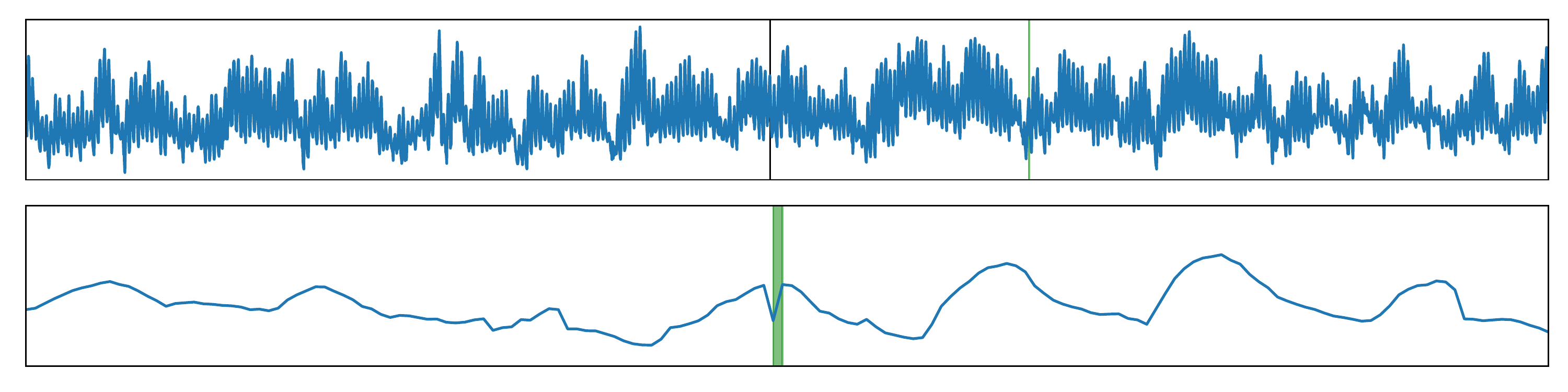} \\ \hline
 Local Peak        & A peak was added, which is lower than the maximal value of the time series.                               & \includegraphics[width=\linewidth]{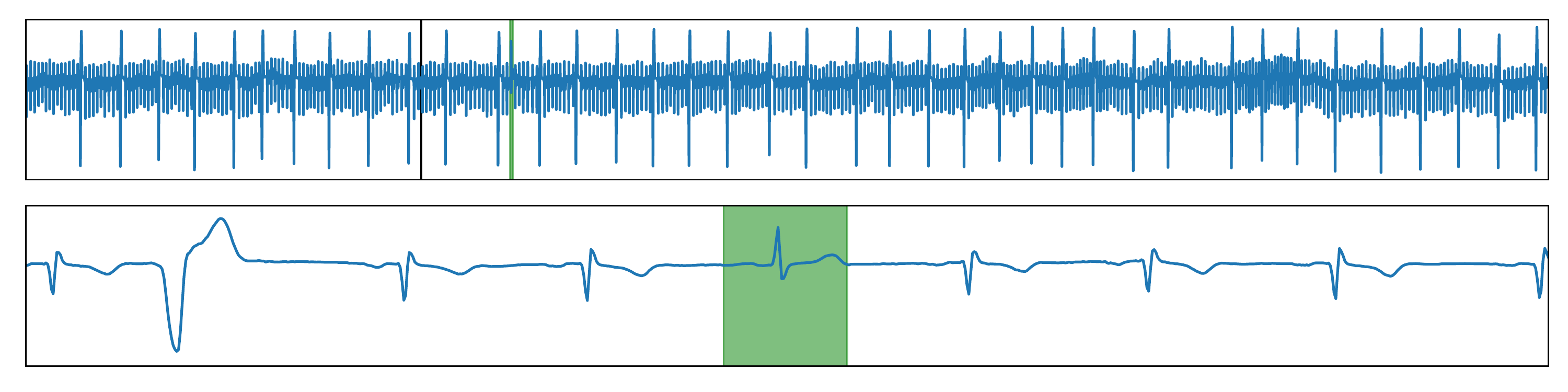}                     \\ \hline
 Missing Drop      & A drop was removed.                                                                                       & \includegraphics[width=\linewidth]{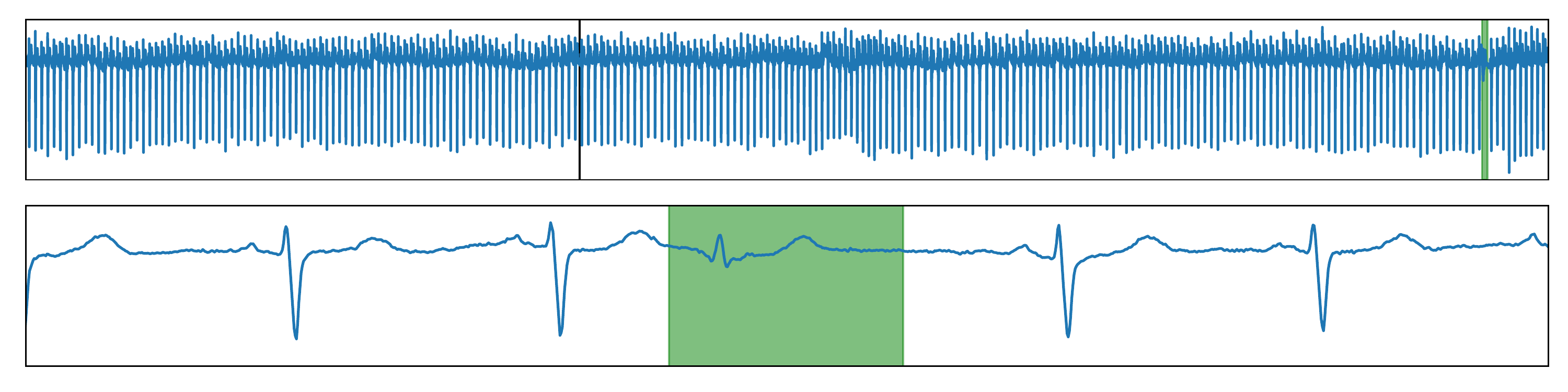}      \\ \hline
 Missing Peak      & A peak was removed.                                                                                       & \includegraphics[width=\linewidth]{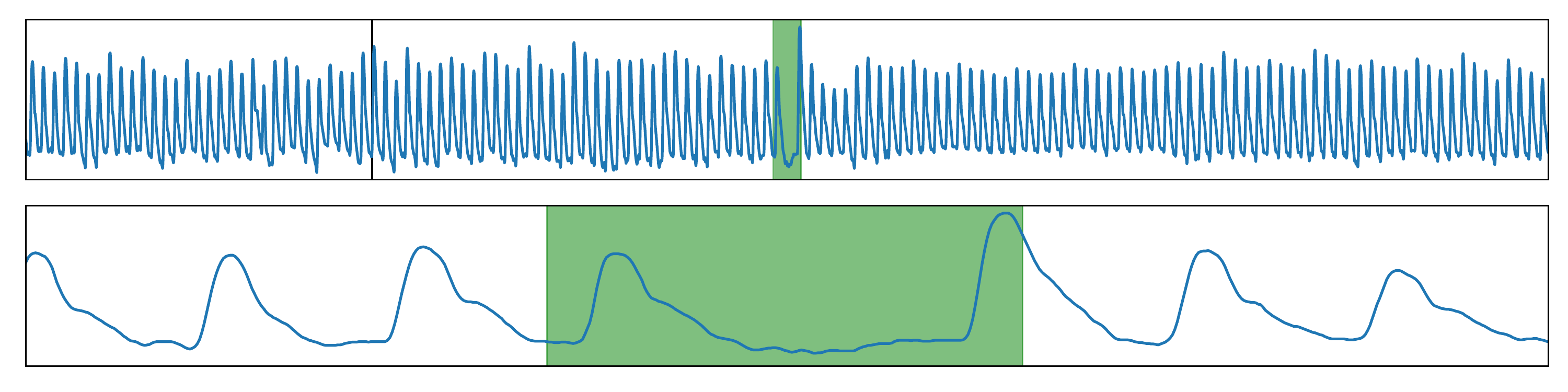}                          \\ \hline
 Noise             & Noise was added to a section.                                                                             & \includegraphics[width=\linewidth]{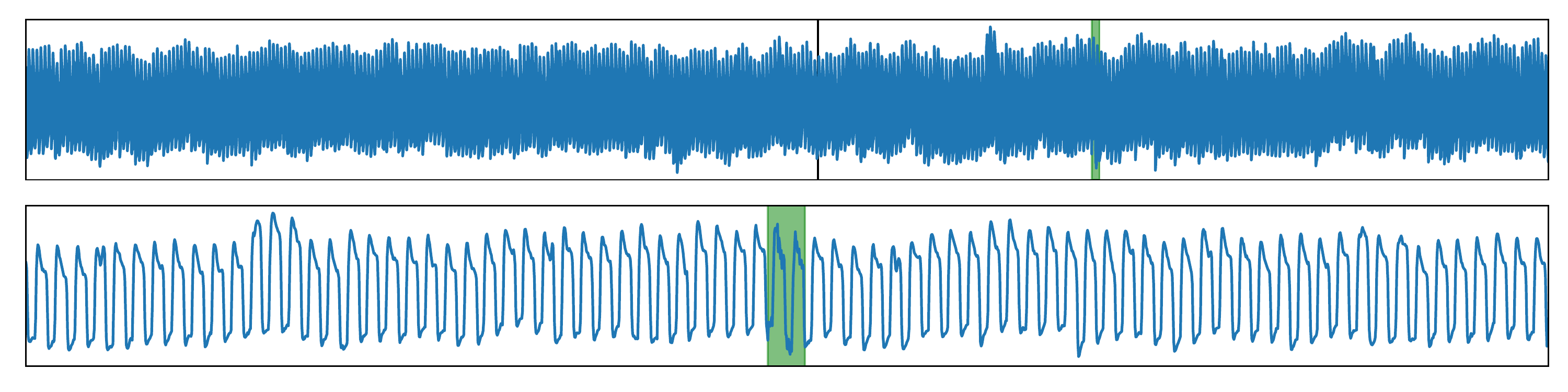}             \\ \hline
 Outlier           & A global outlier.                                                                                         & \includegraphics[width=\linewidth]{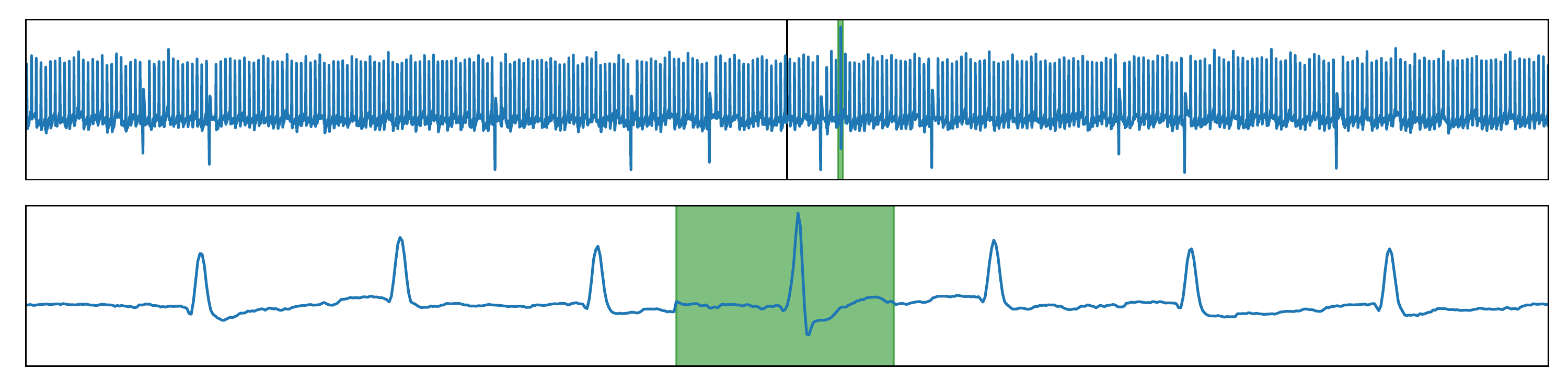}                \\ \hline
 Reversed          & Cycle(s) got reversed.                                                                                    & \includegraphics[width=\linewidth]{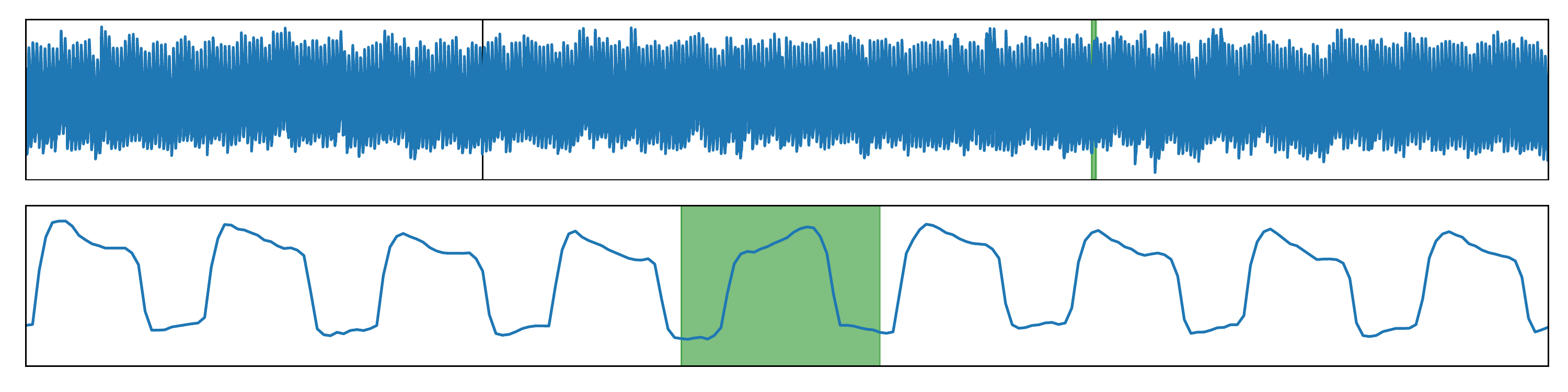}             \\ \hline
 Sampling Rate     & The sampling rate of the signal was increased or decreased in a section.                                  & \includegraphics[width=\linewidth]{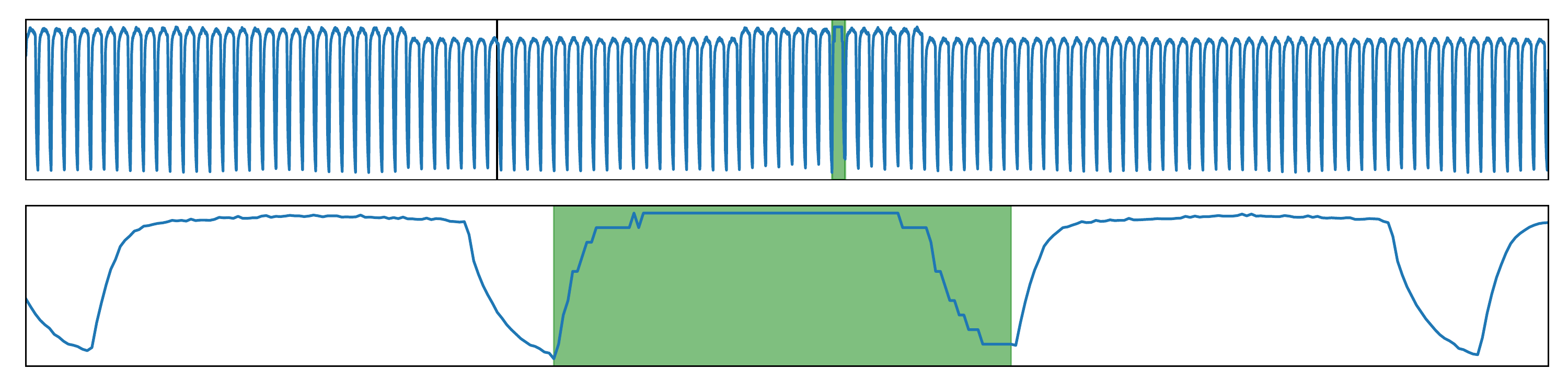}                  \\ \hline
 Signal Shift      & A section was shifted up or down.                                                                         & \includegraphics[width=\linewidth]{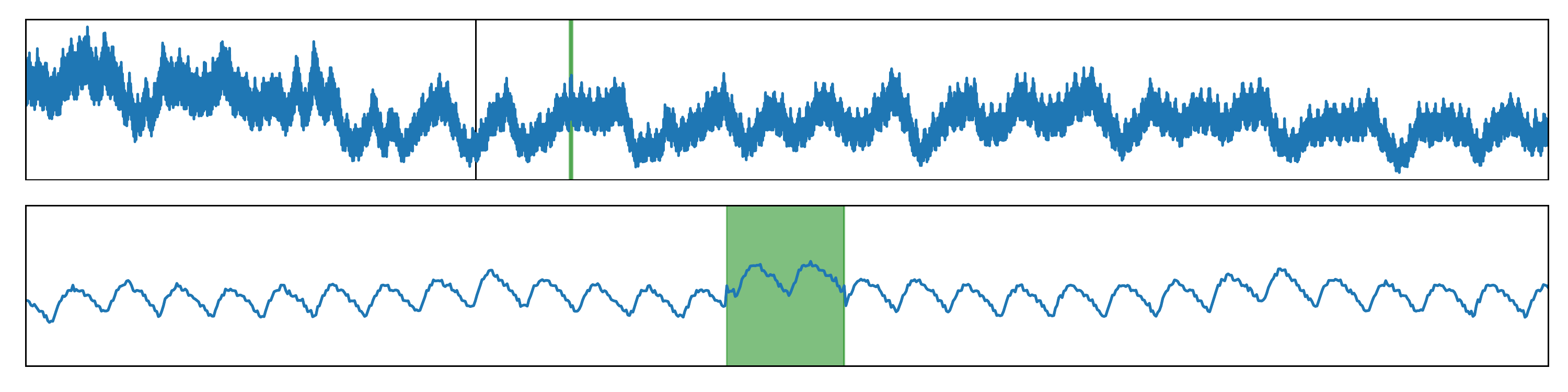}                   \\ \hline
 Smoothed Increase & A otherwise steep increase was smoothed, increasing the number of individual values in this section.      & \includegraphics[width=\linewidth]{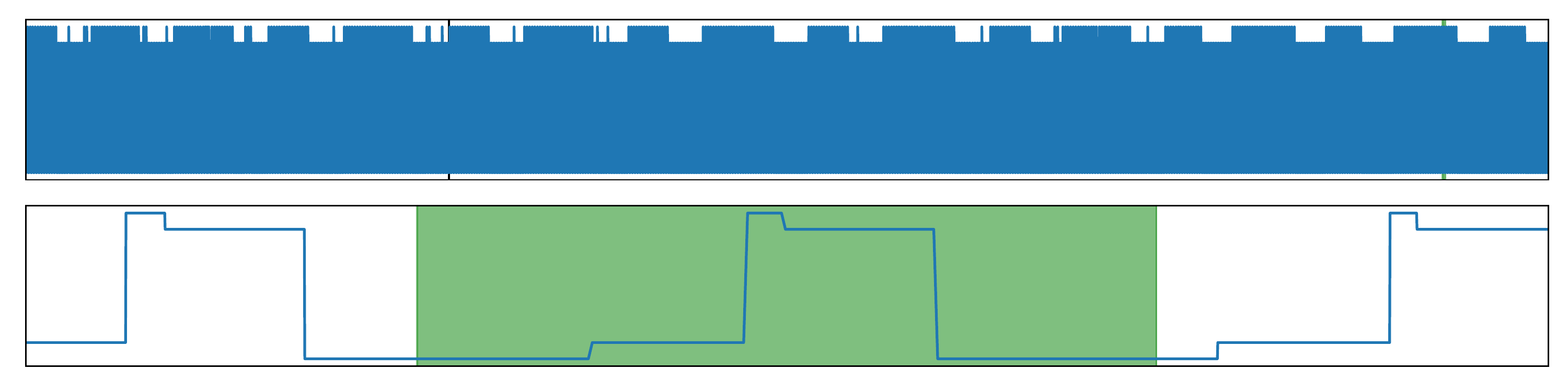}       \\ \hline
 Steep Increase    & A otherwise smooth increase was made steep, reducing the number of individual values within this section. & \includegraphics[width=\linewidth]{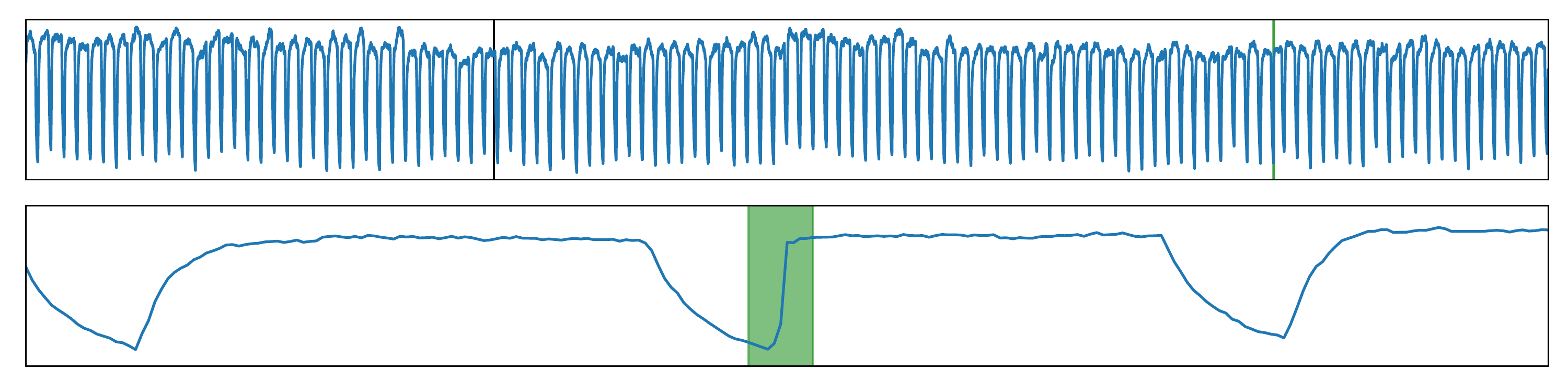}        \\ \hline
 Time Shift        & Increasing the pause between two peaks.                                                                   & \includegraphics[width=\linewidth]{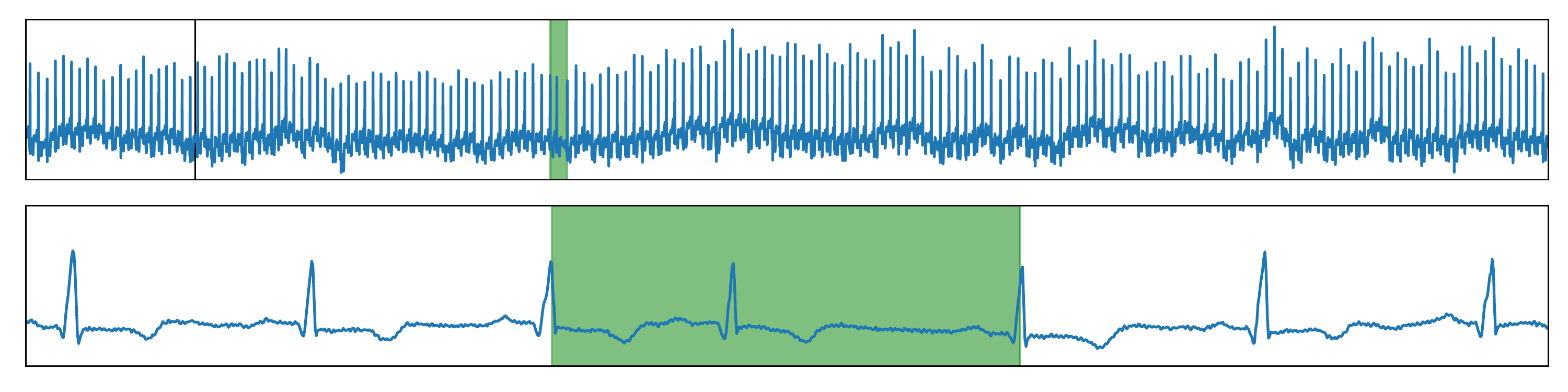}         \\ \hline
 Time Warping      & Moving the cycle peak without changing the cycle length.                                                  & \includegraphics[width=\linewidth]{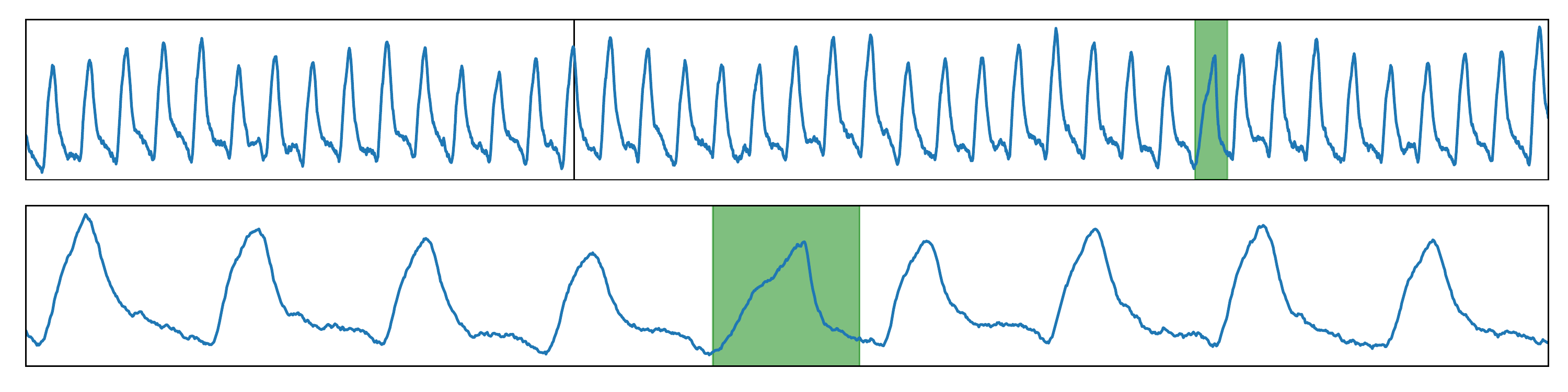}     \\ \hline
 Unusual Pattern   & Replacement of one or more cycle(s) with a different pattern.                                             & \includegraphics[width=\linewidth]{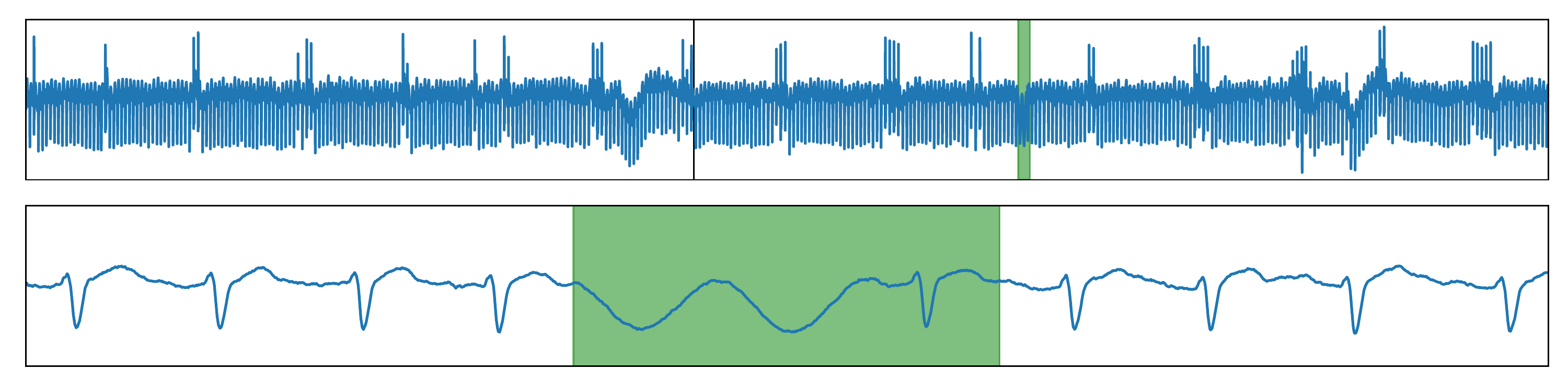}             \\ \hline
\hline
\caption{Description of the annotated anomaly types in the UCR Anomaly Archive dataset.}
\label{tab:appendix:anomaly_types}
\end{longtable}

\subsection{}
\label{appendix:parameters}
\begin{table}[H]
\setlength{\cellWidtha}{\textwidth/4-2\tabcolsep+0in}
\setlength{\cellWidthb}{\textwidth/4-2\tabcolsep+0.4in}
\setlength{\cellWidthc}{\textwidth/4-2\tabcolsep-0.2in}
\setlength{\cellWidthd}{\textwidth/4-2\tabcolsep-0.2in}
\scalebox{1}[1]{\begin{tabularx}{\textwidth}{>{\raggedright\arraybackslash}m{\cellWidtha}>{\raggedright\arraybackslash}m{\cellWidthb}>{\raggedright\arraybackslash}m{\cellWidthc}>{\raggedright\arraybackslash}m{\cellWidthd}}
\toprule
\textbf{}               & \textbf{Parameter}     & \textbf{Value}      & \textbf{Tuned?} \\ \midrule
\textbf{AE}             & subsequence length     & 10                  & no              \\
                        & stride                 & 10                  & no              \\
                        & epochs                 & 20                  & no              \\
                        & batch size             & 32                  & no              \\
                        & latent space dimension & 16                  & yes             \\
                        & learning rate          & 0.005               & yes             \\
                        & weight decay           & $10^{-5}$           & no              \\ \midrule  

\textbf{GANF}           & subsequence length     & 100                 & no              \\
                        & stride                 & 10                  & no              \\
                        & epochs                 & 20 + 30             & no              \\
                        & batch size             & 32                  & no              \\
                        & latent space dimension & 16                  & yes             \\
                        & learning rate          & 0.003               & yes             \\
                        & n\_blocks              & 4                   & yes             \\
                        & weight decay           & $10^{-5}$           & no              \\
                        & h\_tol                 & $10^{-4}$             & no             \\ 
                        & rho\_init              & 1.0                 & no             \\ 
                        & rho\_max                & $10^16$             & no             \\ 
                        & lambda1                & 0.0                 & no             \\ 
                        & alpha\_init            & 0.0                 & no             \\ \midrule
\textbf{MDI}            & $L_{min}$              & 75                  & no              \\
                        & $L_{max}$              & 125                 & no              \\ \midrule
\textbf{MERLIN}         & $L_{min}$              & 75                  & no              \\
                        & $L_{max}$              & 125                 & no              \\ \midrule
\textbf{RRCF}           & n\_trees               & 51                  & yes             \\
                        & tree\_size             & 1001                & yes             \\ \midrule
\textbf{RRCF@sequences} & subsequence length     & 100                 & no              \\
                        & stride                 & 50                  & no              \\
                        & n\_trees               & 68                  & yes             \\
                        & tree\_size             & 150                 & yes             \\ \midrule
\textbf{TranAD}         & subsequence length     & 10                  & no              \\
                        & stride                 & 1                   & no              \\
                        & epochs                 & 1                   & no              \\
                        & batch size             & 128                 & no              \\
                        & learning rate          & 0.02                & yes             \\
                        & weight decay           & $10^{-5}$           & no              \\
                        & step size              & 3                   & yes             \\
                        & gamma                  & 0.75                & yes             \\ \midrule
\end{tabularx}}
\caption{Tuned and not tuned parameters used in our experiments. All other parameters within the methods have been kept to their default values.}
\label{tab:appendix:parameter}
\end{table}

\subsection{}
\label{appendix:stds_by_anomaly_type}
\begin{figure}[H]
\centering
\includegraphics[width=\linewidth]{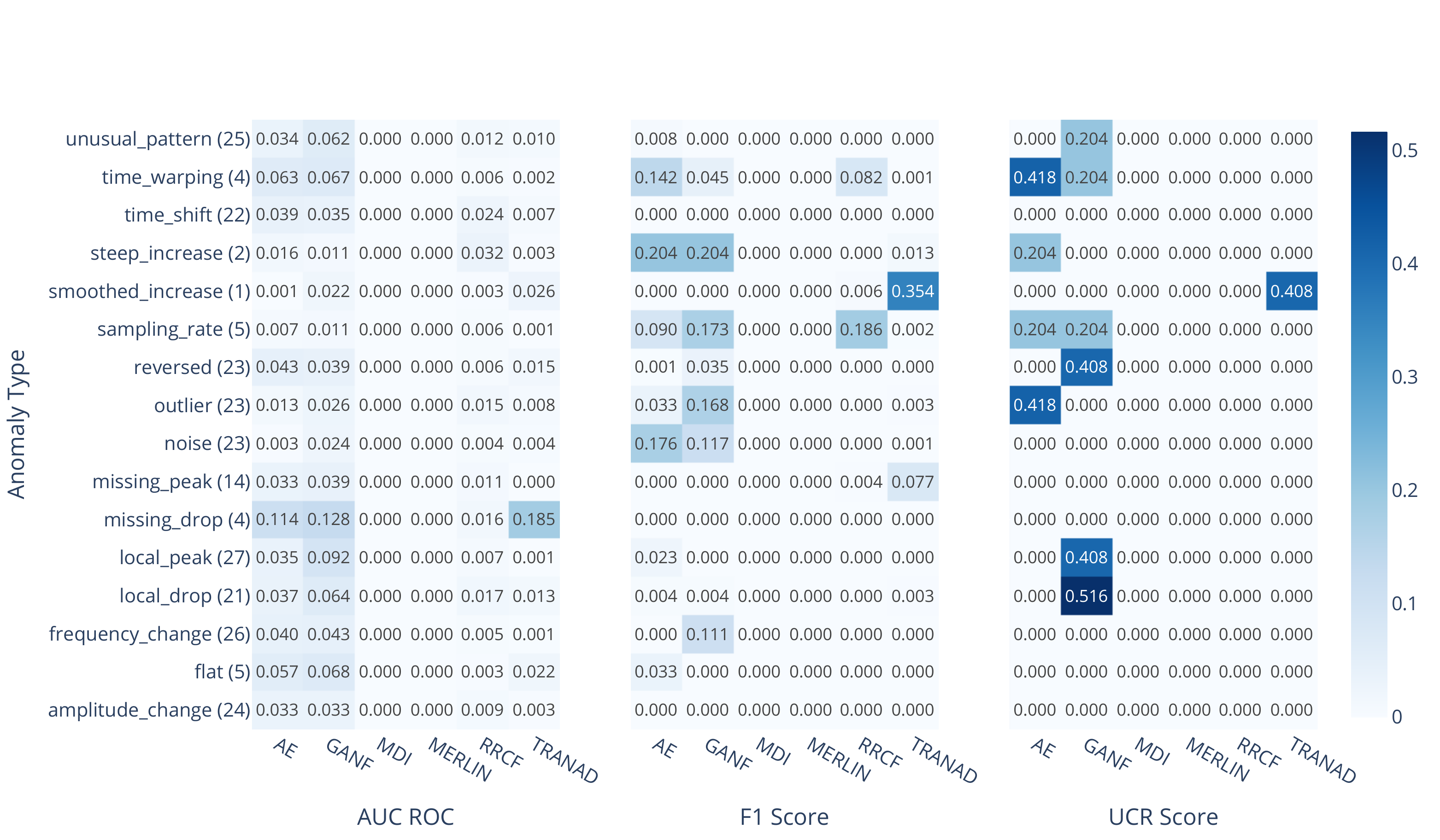}
\caption{The heatmaps show the standard deviation for AUC ROC, F1 Score and UCR Score over six repetitions of the experiment. Next to the anomaly type, the number of times series containing it is shown in parenthesis. The respective macro-averaged mean values can be found in Figure~\ref{fig.results.by_anomaly_type}}
\label{fig.results.std_by_anomaly_type}
\end{figure}   

\end{document}